\documentclass{article}

\usepackage{microtype}
\usepackage{graphicx}
\usepackage{subfigure}
\usepackage{booktabs} 

\usepackage{adjustbox}
\usepackage{diagbox}
\usepackage{times}
\usepackage{epsfig}
\usepackage{amsmath}
\usepackage{amssymb}
\usepackage{colortbl}
\usepackage{stackengine}

\usepackage{multirow}
\usepackage[dvipsnames]{xcolor}
\usepackage{caption}
\usepackage{hyperref}

\definecolor{customgreen}{RGB}{12, 180, 88}
\definecolor{F7E0D5}{RGB}{247,224,213}
\colorlet{Light}{White!0!F7E0D5}

\usepackage[accepted]{icml2025}

\usepackage{amsmath}
\usepackage{amssymb}
\usepackage{mathtools}
\usepackage{amsthm}

\usepackage[capitalize,noabbrev]{cleveref}

\theoremstyle{plain}

\theoremstyle{definition}

\theoremstyle{remark}

\usepackage[textsize=tiny]{todonotes}

\icmltitlerunning{Ranked Entropy Minimization for Continual Test-Time Adaptation}

\begin{document}

\twocolumn[
\icmltitle{Ranked Entropy Minimization for Continual Test-Time Adaptation}




\begin{icmlauthorlist}
\icmlauthor{Jisu Han}{ajou}
\icmlauthor{Jaemin Na}{kt}
\icmlauthor{Wonjun Hwang}{korea}
\end{icmlauthorlist}

\icmlaffiliation{ajou}{Ajou University}
\icmlaffiliation{kt}{Korea Telecom}
\icmlaffiliation{korea}{Korea University}

\icmlcorrespondingauthor{Wonjun Hwang}{wjhwang@korea.ac.kr}

\icmlkeywords{Test-Time Adaptation, Continual Test-Time Adaptation, ICML}

\vskip 0.3in
]



\printAffiliationsAndNotice{} 

\begin{abstract}
Test-time adaptation aims to adapt to realistic environments in an online manner by learning during test time.  Entropy minimization has emerged as a principal strategy for test-time adaptation due to its efficiency and adaptability. Nevertheless, it remains underexplored in continual test-time adaptation, where stability is more important. We observe that the entropy minimization method often suffers from model collapse, where the model converges to predicting a single class for all images due to a trivial solution. We propose ranked entropy minimization to mitigate the stability problem of the entropy minimization method and extend its applicability to continuous scenarios. Our approach explicitly structures the prediction difficulty through a progressive masking strategy. Specifically, it gradually aligns the model's probability distributions across different levels of prediction difficulty while preserving the rank order of entropy. The proposed method is extensively evaluated across various benchmarks, demonstrating its effectiveness through empirical results. Our code is available at \url{https://github.com/pilsHan/rem}
\end{abstract}

\section{Introduction}

\begin{figure}[t]
\begin{center}
\includegraphics[width=0.93\linewidth]{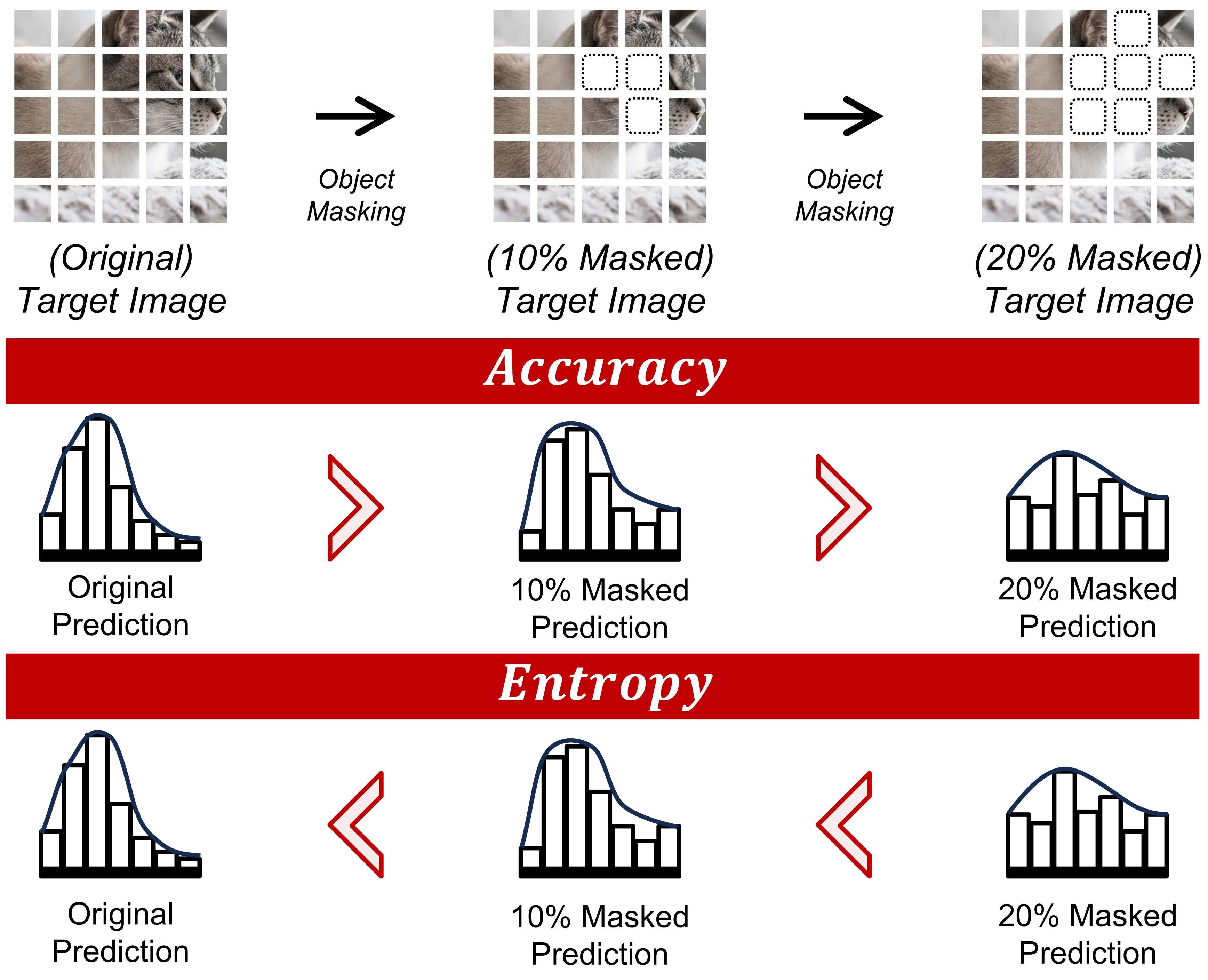}
\end{center}
\vspace{-3mm}
\caption{\textbf{Our Intuition.} We explicitly raise the prediction difficulty of the input images through the masking strategy. Based on the intuition that increased difficulty decreases prediction accuracy and increases entropy, we attempt to maintain a rank ordering of entropy while improving consistency from original to masked predictions. Our approach addresses the problem of model collapse in entropy minimization methods in a simple yet efficient way.}
\label{fig:intuition}
\vspace{-1mm}
\end{figure}

\begin{figure*}[t]
\begin{center}
\vspace{-3mm}
\includegraphics[width=0.95\linewidth]{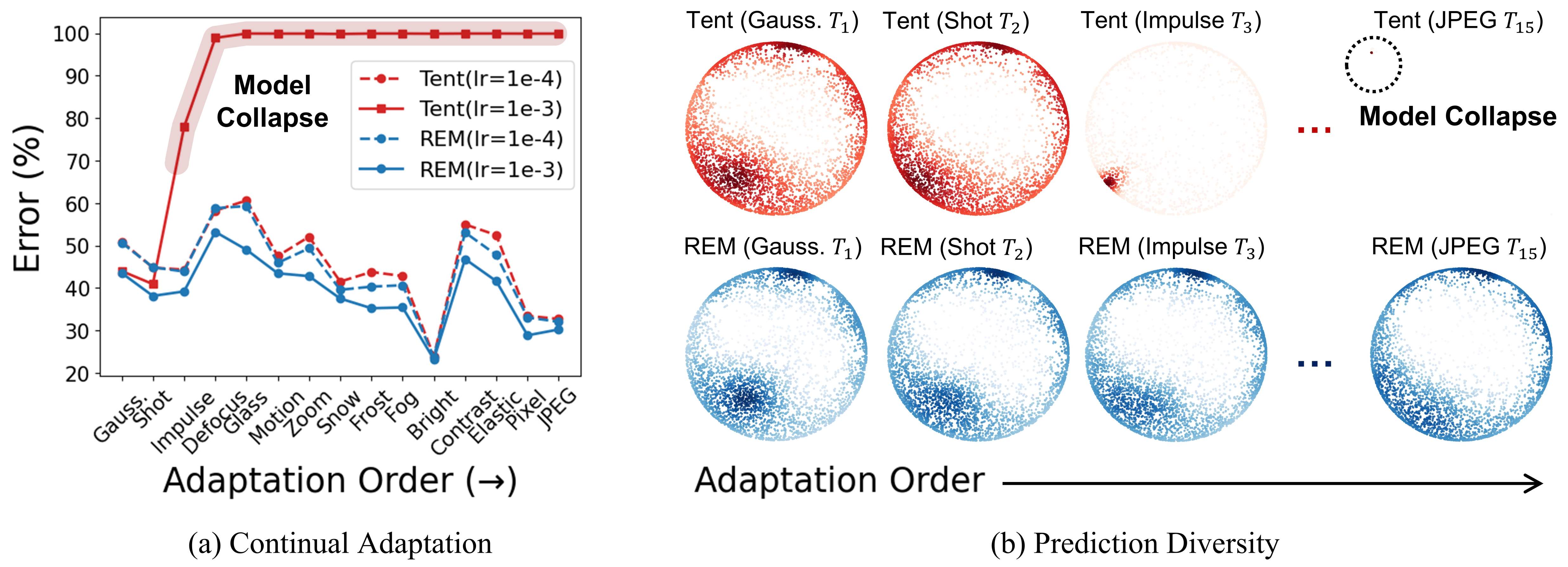}
\end{center}
\caption{\textbf{Observation on model collapse in the entropy minimization approach.} (a) Under the CTTA scenario, the EM approach (\textcolor{BrickRed}{\textbf{Tent}}) undergoes significant performance degradation at a critical point (adaptation order $T_3$, Impulse noise). (b) This phenomenon occurs because the model learns constant representations that do not depend on input images, leading to a collapse in prediction diversity. This is evidenced by class probabilities converging to a single point when visualized in a polar coordinate system. Our proposed method (\textcolor{Blue}{\textbf{REM}}) mitigates model collapse and maintains prediction diversity.}
\label{fig:collapse}
\vspace{-1mm}
\end{figure*}

The real world is non-i.i.d., which demands real-time adaptation of AI applications. Deep learning models have achieved remarkable progress in recent years; however, performance degradation caused by distribution shifts between different domains limits the generalization capabilities~\cite{domainshift}. Test-time adaptation (TTA)~\cite{DequanWangetal2021} has emerged as a practical approach to address non-stationary environmental changes by enabling models trained on source domain to adapt in an online manner to unlabeled target data during the test-time.

Continual test-time adaptation (CTTA)~\cite{Wangetal2022cotta} addresses the issue of error accumulation in long-sequence domains. It mitigates the forgetting problem under continuous environmental changes and sequentially adapts to a stream of data, facilitating the practical deployment of TTA. Recent studies on CTTA can be broadly categorized into two major approaches: entropy minimization (EM)~\cite{DequanWangetal2021,niu2022eata,zhang2024come} and consistency regularization (CR)~\cite{Wangetal2022cotta,liu2023vida,liu2024continual}. The EM approach minimizes the entropy of predictions and offers computational efficiency but suffers from instability due to the risk of trivial solutions, where predictions collapse into a single class. In contrast, the CR approach employs a teacher-student framework~\cite{tarvainen2017meanteacher}, updating the model conservatively to ensure stability but incurs high computational costs. As a result, there is a trade-off between efficiency and stability in these approaches.

\textbf{Motivation.}
Our approach stems from the intuitive observation that a model's predictions for an explicitly information-degraded image are inaccurate and have high entropy compared to its predictions for a complete image (as illustrated in~\cref{fig:intuition}). 
Motivated by Zeno’s Achilles and the tortoise paradox~\cite{huggett2002zeno}, this paper aims to prevent abrupt entropy reduction by gradually advancing the original predictions while reducing the gaps in explicit rank relationships. 
Here, an original prediction is represented by the tortoise, maintaining its leading position over masked prediction in the ranking relationship and learning through rank regularization. In contrast, masked prediction is analogous to Achilles, aiming to catch up with original prediction by learning through consistency regularization.

\textbf{Observation.}~\cref{fig:collapse} shows the phenomenon of model collapse in EM. Model collapse means that the EM approach converges to a prediction for a constant class by a trivial solution, resulting in the model to produce a nonsensical prediction. This issue corresponds to mode collapse in GAN~\cite{goodfellow2014gan} and complete collapse~\cite{hua2021sslcollapse} in self-supervised learning (SSL) within the CTTA. Model collapse occurs because the EM objective function is minimized even when the model consistently predicts a single class, regardless of the input. Since EM lacks stability to address the forgetting problem, it maintains performance by using a low learning rate. However, this leads to limitations in both adaptability and robustness.

Based on this observation, we propose a \textbf{Ranked Entropy Minimization (REM)}. Specifically, we exploit the self-attention structure of ViT~\cite{dosovitskiy2021vit} to mask patches with a high likelihood of containing objects~\cite{bolya2023token,son2023maskedkd}. The principal idea is to explicitly enhance the prediction complexity of a sample by masking objects that domain invariant features. Building a mask chain that sequentially obscures more patches based on the masking ratio transforms unpredictable prediction tendencies into a ranked predictable one. For taking advantage of the ranked structure, we provide two interrelated methods. (1) First, we apply a consistency loss by ensuring that predictions with a higher masking ratio are similar to those with a lower masking ratio, thereby not only indirectly reducing prediction entropy but also enabling the model to learn contextual information from masked regions. (2) Second, we introduce a ranking loss that ensures the entropy of predictions with a lower masking ratio remains lower than that of predictions with a higher masking ratio. This approach not only models uncertainty but also reduces entropy by incorporating object-specific information, preventing predictions from being biased toward background information. The main benefit of our method lies in achieving the joint goals of stability and adaptability while maintaining the efficiency of EM approaches with a single model and without requiring additional models.

\section{Related Work}
\subsection{Test-Time Adaptation}
The concept of optimizing during test time to adapt to target domains was proposed in Test-Time Training (TTT)~\cite{sun2020ttt}. However, TTT relies on the use of source data and training loss, which may not generalize well to practical applications. To address these limitations, Tent~\cite{DequanWangetal2021} proposed the concept of Fully TTA, which enables online adaptation to unlabeled target data without requiring access to source data. In this context, TTA focuses on efficiency, typically by updating the normalization layers~\cite{gong2022note} or adjusting predictions without training the model parameters to improve computational efficiency.~\cite{iwasawa2021t3a,boudiaf2022lame}.

Entropy minimization approaches~\cite{DequanWangetal2021,niu2022eata,niu2023sar,lee2024deyo,zhang2024come} are evolving as a primary solution for TTA. Among those, EATA~\cite{niu2022eata} introduces sample filtering for uncertain predictions and regularization method to enhance stability. Furthermore, SAR~\cite{niu2023sar} observes the phenomenon of model collapse by trivial solutions, and proposes adaptation to flat minima and filtering of large gradient samples. Building on these approaches, DeYO~\cite{lee2024deyo} presents a criteria for additional sample selection through image rearrangement, and COME~\cite{zhang2024come} proposes a conservative entropy minimization method to address the overconfidence problem.

\subsection{Continual Test-Time Adaptation}
Advancing from single-domain adaptation, CTTA~\cite{Wangetal2022cotta,brahma2023petal,dobler2023rmt,liu2023vida,liu2024continual} emphasizes the requirement for adapting to continuous domain shifts, thereby extending the practicality of TTA. This paradigm promotes rethinking the conventional protocol by encouraging a deeper focus on stability. CoTTA~\cite{Wangetal2022cotta} addresses catastrophic forgetting by introducing a consistency loss between the base model and weight-averaged model, while employing stochastic restoration of parameters based on the source model. PETAL~\cite{brahma2023petal} introduces a probabilistic framework for CTTA and a parameter restoration method leveraging the Fisher Information Matrix. ViDA~\cite{liu2023vida} presents a trade-off between stability and plasticity by designing an adapter that explicitly separates domain-invariant features and domain-specific features. Continual-MAE~\cite{liu2024continual} measures pixel uncertainty through Monte Carlo (MC) dropout to distinguish object presence and enhances the representation of domain-invariant properties using a masked autoencoder.

\textbf{Beyond not Forgetting.} The recent state-of-the-art CTTA methods achieve stability and mitigate performance degradation by adopting teacher-student frameworks or parameter restoration from the source model. However, these approaches often neglect efficiency constraints, leading to excessive computational costs and memory overhead. This inefficiency contradicts the goal of TTA, which is to enable real-time adaptation under resource constraints. To address this issue and realign the framework with its intended purpose, protocols that account for computational time constraints~\cite{alfarra2024constraints} and label delays~\cite{csaba2024labeldelay} have been proposed. We argue that efficiency in CTTA requires renewed attention and propose leveraging insights from pioneering approaches in TTA to improve both stability and computational efficiency. In this study, we integrate the strengths of EM and CR approaches to balance efficiency and stability. Detailed efficiency analysis is provided in~\cref{Efficiency Comparison}.

\section{Ranked Entropy Minimization}
In this section, we introduce the CTTA setup and the trivial solution from the EM, then explain the proposed REM. Our method consists of a mask chaining strategy and two loss functions for CR and EM. Masked consistency loss ensures consistency from predictions with a lower masking ratio to those with a higher masking ratio, while entropy ranking loss enforces a ranking constraint from predictions with a higher masking ratio to those with a lower masking ratio.

\subsection{Preliminaries}
In CTTA setup, a target model $f_t$ is trained in an online manner, starting from a source model $f_s$. The model adapts to sequential target domain data $x_t^{\tau}$ , where each domain is represented in order by $\tau \in \{1, 2, \dots\}$. The evaluation protocol involves calculating the cumulative error based on the model's predictions $\hat{p}_t^{\tau} = f_t(x_t^{\tau})$, while the model adapts to the target domains during test time using $x_t^{\tau}$. The main difference between TTA and CTTA lies in whether the model is reset to the source model when a domain changes. In CTTA, the model is not reset, which makes the issue of catastrophic forgetting more significant.

The EM method applies gradient descent using entropy 
$\mathcal{S}(\hat{p}_t) = - \sum^{C}_{c=1} \hat{p}_{t,c}\log \hat{p}_{t,c}$
over the total number of classes $C$ as the objective function. Consequently, the gradient of $\mathcal{S}(\hat{p}_t)$ with respect to the parameter 
$\theta$ is as follows:
\begin{align}
\frac{\partial \mathcal{S}(\hat{p}_t)}{\partial \theta_t} &= - \sum_{i=1}^{C} \left( \log \hat{p}_{t,c} + 1 \right) \hat{p}_{t,c} (1 - \hat{p}_{t,c}) \frac{\partial z_{t,c}}{\partial \theta_t},
\end{align}
where $z_{t,c}$ represents the logit for class $c$ before applying the softmax function, i.e., $\hat{p}_{t,c} = \frac{\exp(z_{t,c})}{\sum_{i=1}^{C} \exp(z_{t,i})}$. The trivial solution for \(\frac{\partial \mathcal{S}(\hat{p}_t)}{\partial \theta_t} = 0\) arises under the following cases:
\textbf{(Case 1)} Uniformly distributed probability $(\hat{p}_{t,c} = \frac{1}{C})$, and \textbf{(Case 2)} Perfectly confident predictions $(\hat{p}_{t,c} \in \{0, 1\})$.
In this case, since the initial source model's prediction distribution is not uniform, the likelihood of achieving a trivial solution due to a uniform distribution under entropy minimization is low. However, it is possible for a singular class to result in perfectly confident predictions, which is experimentally observed in~\cref{fig:collapse}.

\begin{figure}[t]
\begin{center}
\includegraphics[width=\linewidth]{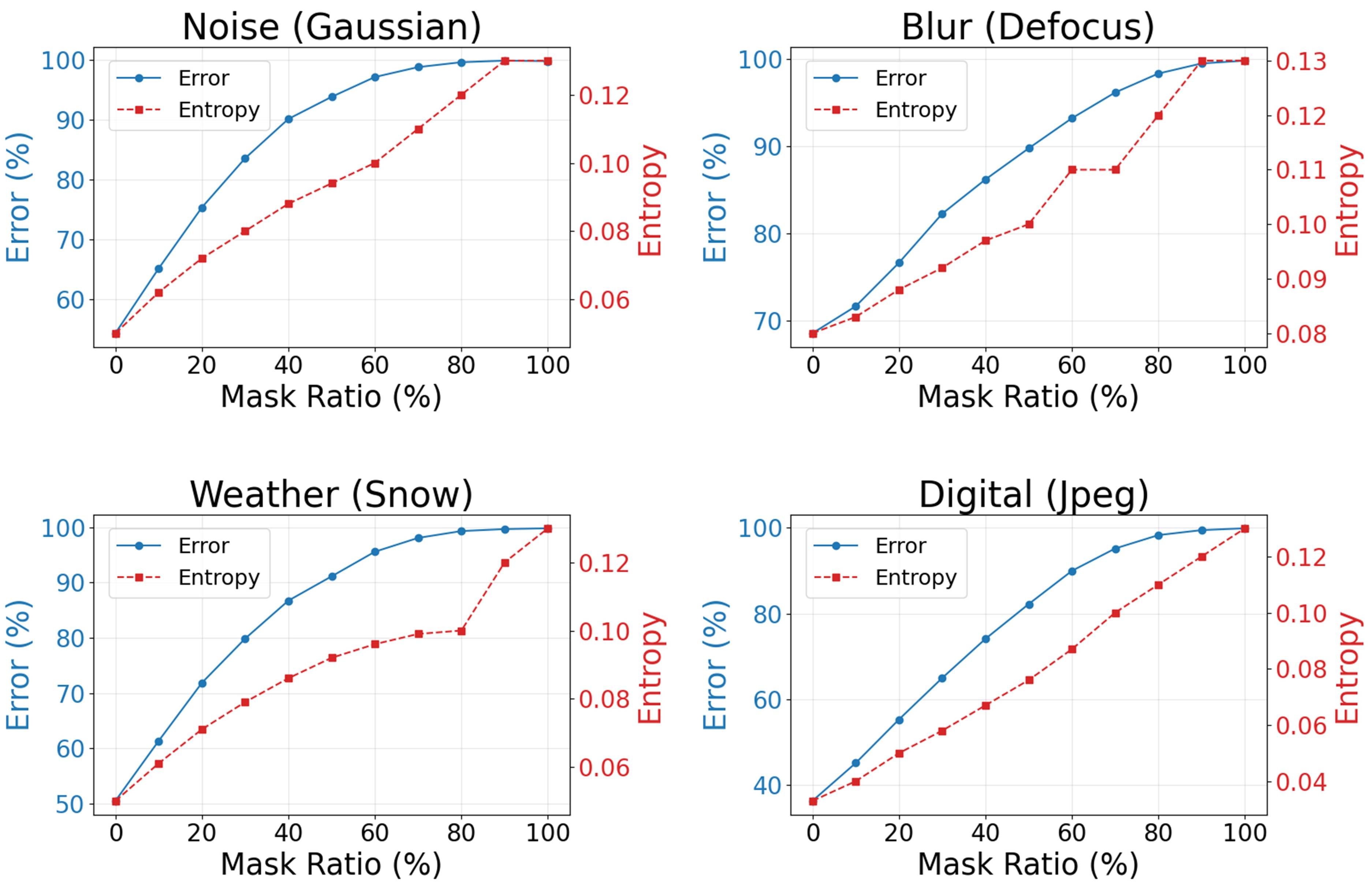}
\end{center}
\caption{\textbf{Empirical study according to masking ratio.} We report the changes in error and entropy as the masking ratio increases. Both entropy and error exhibit a monotone increasing trend with respect to the masking ratio, and we observe that linearity becomes more pronounced, especially in regions with lower masking ratios.}
\label{fig:accandentropy}
\end{figure}

\subsection{Explicit Mask Chaining}
Conventional CTTA methods point out that augmentation policies may not be valid for dramatic domain shifts and decide whether to apply augmentation through prediction confidence~\cite{Wangetal2022cotta}. However, these methods fail to model the impact of augmentation policies on model predictions, resulting in inefficiencies caused by the application of diverse augmentations. In order to model the predictions, we propose Explicit Mask Chaining, which incrementally masks content containing domain-invariant information for augmentation. To this end, we focus on the self-attention structure of ViT for efficient object masking. Exploiting the self-attention structure allows us to cluster content by similarity between intermediate tokens~\cite{bolya2023token}, as well as compute an attention score for an attention head~\cite{son2023maskedkd}. Following pioneering research, we define an attention score $A$ as follows:
\begin{align}
A = \sum_{h=1}^{H} \mathrm{Softmax}\left(\frac{Q_{h,cls} {K_{h,img}}^\top}{\sqrt{d}}\right),
\end{align}
where $H$ is the number of heads in the multi-head attention, $Q_{h,cls}$ and $K_{h,img}$ are query for class token and key for image token of the $h$-th attention head, respectively. $d$ denotes the dimension of each attention head.

The masked image $\{x_{m_1}, x_{m_2}, \cdots, x_{m_N}\}$, defined for the top-$m$ proportion of a set $A$ sorted in descending order, has mask ratios that satisfy the condition $0 \leq m_1 \leq m_2 \leq \cdots \leq m_N \leq 1$.
~\cref{fig:accandentropy} shows the error and entropy across a range of masking ratios for noise, blur, weather, and digital corruptions on ImageNetC~\cite{hendrycks2019benchmarking}. The results empirically confirm that explicit masking of objects results in lower accuracy and higher entropy as the masking ratio increases. In conclusion, we establish an explicit ranking relationship between entropy and accuracy by employing an incremental masking strategy for consecutive mask ratio.

\subsection{Masked Consistency Loss}
Given an explicitly ranked prediction, we aim to make a high masking rate prediction with relatively low accuracy similar to a low mask rate prediction with relatively high accuracy. We define masked consistency loss (MCL) as follows:
\begin{align}
\label{eq:mcl}
\mathcal{L}_{MCL} = \sum_{i<j}^{M_N} \mathcal{H}(f_t(x_{j}),\mathbf{sg}(f_t(x_{i}))), 
\end{align}
where $\mathcal{H}(p,q)$ denotes the cross-entropy between two probability distributions $p$ and $q$, $M_N = \{0, m_1, m_2, \dots, m_N\}$ is a set of mask ratios, where $N$ is the number of mask chains, and $\mathbf{sg}$ denotes the stop-gradient operation.

\noindent\textbf{Comparison with EM.} Unlike EM, which uses entropy as a loss function, MCL indirectly reduces prediction entropy by using the cross-entropy between masked predictions and either less-masked or unmasked predictions. This is designed to mitigate abrupt entropy changes and overconfident predictions, leading to the alleviation of model collapse.

\noindent\textbf{Comparison with CR.} To generate stable predictions distinct from the target model, the CR approach requires additional teacher models and numerous forward passes for uncertainty estimation. In contrast, our method eliminates the need for uncertainty prediction by leveraging an explicit ranking structure and improves efficiency by generating diverse predictions via mask chains within a single model.

\subsection{Entropy Ranking Loss}
Through the reduction of differences in ranked prediction distributions, MCL is designed to indirectly minimize prediction entropy. However, this may result in slower adaptation due to small differences in prediction distributions or lead to biased predictions toward the background when learning from images with occluded objects. To complement MCL and address these issues, we propose the entropy ranking loss (ERL) as follows:
\begin{align}
\label{eq:erl}
\mathcal{L}_{ERL} = \sum_{i<j}^{M_N} \max{(0, S(f_t(x_i))-\mathbf{sg}(S(f_t(x_j)))+\mathsf{m})},
\end{align}
where $\mathrm{m}$ is the margin. The purpose of ERL is to maintain the principle that the entropy of predictions with a low masking ratio for objects should be lower than that of predictions with a high masking ratio. By maintaining an explicit ranked order, we prevent overconfidence in high masking ratio predictions, which could lead to bias toward the background. This follows from prior findings that applying ranking losses in neural calibration effectively mitigates overconfidence~\cite{moon2020confidence,noh2023rankmixup}. Additionally, ERL directly reduces the entropy of samples that violate the ranked order, thereby promoting faster adaptation and maintaining a structured entropy hierarchy across different masking ratios.

\begin{table*}[t]
\centering
\caption{\label{tab:imagenet}Classification error rate (\%) for ImageNet-to-ImageNetC under CTTA scenario. Mean (\%) denotes the average error rate across 15 target domains. Gain (\%) represents the relative performance improvement compared to the source model.}
\small
\setlength\tabcolsep{2pt}
\begin{adjustbox}{width=1\linewidth,center=\linewidth}
\begin{tabular}{l|ccccccccccccccc|cc}
\hline
Time & \multicolumn{15}{l|}{$t\xrightarrow{\hspace*{13.5cm}}$}& \\ \hline
Method &
\rotatebox[origin=c]{50}{Gaussian} & \rotatebox[origin=c]{50}{shot} & \rotatebox[origin=c]{50}{impulse} & \rotatebox[origin=c]{50}{defocus} & \rotatebox[origin=c]{50}{glass} & \rotatebox[origin=c]{50}{motion} & \rotatebox[origin=c]{50}{zoom} & \rotatebox[origin=c]{50}{snow} & \rotatebox[origin=c]{50}{frost} & \rotatebox[origin=c]{50}{fog}  & \rotatebox[origin=c]{50}{brightness} & \rotatebox[origin=c]{50}{contrast} & \rotatebox[origin=c]{50}{elastic\_trans} & \rotatebox[origin=c]{50}{pixelate} & \rotatebox[origin=c]{50}{jpeg}
& Mean$\downarrow$ & Gain\\\hline
Source~\cite{dosovitskiy2021vit}&53.0&51.8&52.1&68.5&78.8&58.5&63.3&49.9&54.2&57.7&26.4&91.4&57.5&38.0&36.2&55.8&0.0\\
Pseudo-label~\cite{Leeetal2013}&45.2&40.4&41.6&51.3&53.9&45.6&47.7&40.4&45.7&93.8&98.5&99.9&99.9&98.9&99.6&61.2&-5.4\\
Tent~\cite{DequanWangetal2021}&52.2&48.9&49.2&65.8&73.0&54.5&58.4&44.0&47.7&50.3&23.9&72.8&55.7&34.4&33.9&51.0&+4.8\\
CoTTA~\cite{Wangetal2022cotta}&52.9&51.6&51.4&68.3&78.1&57.1&62.0&48.2&52.7&55.3&25.9&90.0&56.4&36.4&35.2&54.8&+1.0\\
VDP~\cite{gan2023decorate} &52.7&51.6&50.1&58.1&70.2&56.1&58.1&42.1&46.1&45.8&23.6&70.4&54.9&34.5&36.1&50.0&+5.8\\
SAR~\cite{niu2023sar}&49.3&43.8&44.9&58.2&60.9&46.1&51.8&41.3&44.1&41.8&23.8&57.2&49.9&32.9&32.7&45.2&+10.6\\
PETAL~\cite{brahma2023petal}&52.1&48.2&47.5&66.8&74.0&56.7&59.7&46.8&47.2&52.7&26.4&91.3&50.7&32.3&32.0&52.3&+3.5\\
ViDA~\cite{liu2023vida}&47.7&42.5 &42.9& 52.2 & 56.9 & 45.5 & 48.9 & 38.9 & 42.7 & 40.7 & 24.3 & 52.8 & 49.1 & 33.5 & 33.1 & 43.4 & +12.4 \\
Continual-MAE~\cite{liu2024continual} &46.3&41.9&42.5&\bf51.4&54.9&\bf43.3&\bf40.7&\bf34.2&35.8&64.3&23.4&60.3&\bf37.5&29.2&31.4&42.5& +13.3\\
\rowcolor{Light!70}REM (Ours)
&\bf43.5&\bf38.1&\bf39.2&53.2&\bf49.0&43.5&42.8&37.5&\bf35.2&\bf35.4&\bf23.2&\bf46.8&41.6&\bf28.9&\bf30.2&\bf39.2&\bf+16.6\\
\midrule
\rowcolor{gray!10}Supervised &42.5&36.9&37.1&46.4&44.0&37.4&38.3&34.2&33.1&32.5&21.5&43.3&34.4&26.1&27.5&35.7&+20.1\\
\hline
\end{tabular}
\end{adjustbox}
\vspace{-2mm}
\end{table*}

\begin{table*}[t]
\centering
\caption{\label{tab:cifar10}Classification error rate (\%) for CIFAR10-to-CIFAR10C under CTTA scenario. Mean (\%) denotes the average error rate across 15 target domains. Gain (\%) represents the relative performance improvement compared to the source model.}
\small
\setlength\tabcolsep{2pt}
\begin{adjustbox}{width=1\linewidth,center=\linewidth}
\begin{tabular}{l|ccccccccccccccc|cc}
\hline
Time & \multicolumn{15}{l|}{$t\xrightarrow{\hspace*{13.5cm}}$}& \\ \hline
Method  &
\rotatebox[origin=c]{50}{Gaussian} & \rotatebox[origin=c]{50}{shot} & \rotatebox[origin=c]{50}{impulse} & \rotatebox[origin=c]{50}{defocus} & \rotatebox[origin=c]{50}{glass} & \rotatebox[origin=c]{50}{motion} & \rotatebox[origin=c]{50}{zoom} & \rotatebox[origin=c]{50}{snow} & \rotatebox[origin=c]{50}{frost} & \rotatebox[origin=c]{50}{fog}  & \rotatebox[origin=c]{50}{brightness} & \rotatebox[origin=c]{50}{contrast} & \rotatebox[origin=c]{50}{elastic\_trans} & \rotatebox[origin=c]{50}{pixelate} & \rotatebox[origin=c]{50}{jpeg}
& Mean$\downarrow$ & Gain\\\hline
Source \cite{dosovitskiy2021vit}&60.1&53.2&38.3&19.9&35.5&22.6&18.6&12.1&12.7&22.8&5.3&49.7&23.6&24.7&23.1&28.2&0.0\\
Pseudo-label \cite{Leeetal2013}&59.8&52.5&37.2&19.8&35.2&21.8&17.6&11.6&12.3&20.7&5.0&41.7&21.5&25.2&22.1&26.9&+1.3\\
Tent \cite{DequanWangetal2021}&57.7&56.3&29.4&16.2&35.3&16.2&12.4&11.0&11.6&14.9&4.7&22.5&15.9&29.1&19.5&23.5&+4.7\\
CoTTA \cite{Wangetal2022cotta}&58.7&51.3&33.0&20.1&34.8&20&15.2&11.1&11.3&18.5&4.0&34.7&18.8&19.0&17.9&24.6&+3.6\\
VDP \cite{gan2023decorate} &57.5&49.5&31.7&21.3&35.1&19.6&15.1&10.8&10.3&18.1&4.0&27.5&18.4&22.5&19.9&24.1&+4.1\\
SAR~\cite{niu2023sar}&54.1&47.6&38.0&19.9&34.8&22.6&18.6&12.1&12.7&22.8&5.3&39.9&23.6&24.7&23.1&26.6&+1.6\\
PETAL~\cite{brahma2023petal}&59.9&52.3&36.1&20.1&34.7&19.4&14.8&11.5&11.2&17.8&4.4&29.6&17.6&19.2&17.3&24.4&+3.8\\
ViDA \cite{liu2023vida}& 52.9&47.9&19.4&11.4&31.3&13.3&7.6&7.6&9.9&12.5&3.8&26.3&14.4&33.9&18.2&20.7&+7.5 \\
Continual-MAE~\cite{liu2024continual}& 30.6 & 18.9 & 11.5 & 10.4 & 22.5 & 13.9 & 9.8 & \textbf{6.6} & 6.5 & 8.8 & 4.0 & 8.5 & 12.7 & 9.2 & 14.4 &12.6& +15.6\\
\rowcolor{Light!70}REM (Ours)
&\bf17.3&\bf12.5&\bf10.3&\bf8.4&\bf17.7&\bf8.4&\bf5.5&\bf6.6&\bf5.6&\bf7.2&\bf3.7&\bf6.4&\bf11.0&\bf7.3&\bf13.0&\bf9.4&\bf+18.8\\
\midrule
\rowcolor{gray!10}Supervised &14.6&9.0&6.9&6.1&11.2&6.0&3.7&4.4&3.4&4.9&2.1&3.7&7.5&4.3&8.5&6.4&+21.8\\
\hline
\end{tabular}
\end{adjustbox}
\vspace{-0.15cm}
\end{table*}

\begin{table*}[t]
\caption{\label{tab:cifar100}Classification error rate (\%) for CIFAR100-to-CIFAR100C under CTTA scenario. Mean (\%) denotes the average error rate across 15 target domains. Gain (\%) represents the relative performance improvement compared to the source model.}
\small
\centering
\setlength\tabcolsep{2pt}
\begin{adjustbox}{width=1\linewidth,center=\linewidth}
\begin{tabular}{l|ccccccccccccccc|cc}
\hline
Time & \multicolumn{15}{l|}{$t\xrightarrow{\hspace*{13.5cm}}$}& \\ \hline
Method &
\rotatebox[origin=c]{50}{Gaussian} & \rotatebox[origin=c]{50}{shot} & \rotatebox[origin=c]{50}{impulse} & \rotatebox[origin=c]{50}{defocus} & \rotatebox[origin=c]{50}{glass} & \rotatebox[origin=c]{50}{motion} & \rotatebox[origin=c]{50}{zoom} & \rotatebox[origin=c]{50}{snow} & \rotatebox[origin=c]{50}{frost} & \rotatebox[origin=c]{50}{fog}  & \rotatebox[origin=c]{50}{brightness} & \rotatebox[origin=c]{50}{contrast} & \rotatebox[origin=c]{50}{elastic\_trans} & \rotatebox[origin=c]{50}{pixelate} & \rotatebox[origin=c]{50}{jpeg}
& Mean$\downarrow$ & Gain\\\hline
Source~\cite{dosovitskiy2021vit}&55.0&51.5&26.9&24.0&60.5&29.0&21.4&21.1&25.0&35.2&11.8&34.8&43.2&56.0&35.9&35.4&0.0\\
Pseudo-label~\cite{Leeetal2013}&53.8&48.9&25.4&23.0&58.7&27.3&19.6&20.6&23.4&31.3&11.8&28.4&39.6&52.3&33.9&33.2&+2.2\\
Tent~\cite{DequanWangetal2021} &53.0&47.0&24.6&22.3&58.5&26.5&19.0&21.0&23.0&30.1&11.8&25.2&39.0&47.1&33.3&32.1&+3.3\\
CoTTA~\cite{Wangetal2022cotta}&55.0&51.3&25.8&24.1&59.2&28.9&21.4&21.0&24.7&34.9&11.7&31.7&40.4&55.7&35.6&34.8&+0.6\\
VDP~\cite{gan2023decorate}&54.8&51.2&25.6&24.2&59.1&28.8&21.2&20.5&23.3&33.8&\textbf{7.5}&\textbf{11.7}&32.0&51.7&35.2&32.0&+3.4\\
SAR~\cite{niu2023sar}&39.4&31.0&19.8&20.9&43.9&22.6&19.1&20.3&20.2&24.3&11.8&22.3&35.2&32.1&30.1&26.2&+9.2\\
PETAL~\cite{brahma2023petal}&49.2&38.7&24.1&26.3&38.2&25.4&19.4&21.0&19.3&26.6&15.4&31.8&28.3&26.6&29.5&28.0&+7.4\\
ViDA~\cite{liu2023vida}&50.1 & 40.7 & 22.0 & 21.2 & 45.2 & 21.6 & \textbf{16.5} & \textbf{17.9} & \textbf{16.6} & 25.6 & 11.5 & 29.0 & \textbf{29.6} & 34.7 & \textbf{27.1} & 27.3 & +8.1\\
Continual-MAE~\cite{liu2024continual}& 48.6 & 30.7 & 18.5 & 21.3 & 38.4 & 22.2 & 17.5 & 19.3 & 18.0 & 24.8 & 13.1 & 27.8 & 31.4 & 35.5 & 29.5&26.4 &+9.0 \\
\rowcolor{Light!70}REM (Ours)
&\bf29.2& \bf25.5& \bf17.0& \bf19.1& \bf35.2& \bf21.2& 18.3& 19.5& 18.7& \bf22.8 &15.5& 17.6& 31.6& \bf26.2& 33.0& \bf23.4&\bf+12.0\\
\midrule
\rowcolor{gray!10}Supervised &26.2&20.6&13.9&15.9&24.6&15.6&11.8&13.1&12.1&13.6&8.5&9.7&20.2&13.5&21.5&16.1&+19.3\\
\hline
\end{tabular}
\end{adjustbox}
\end{table*}

\subsection{Total Loss Function}
The total loss function $\mathcal{L}_{REM}$ is expressed as a linear combination of $\mathcal{L}_{MCL}$ in Eq.~\ref{eq:mcl} and $\mathcal{L}_{ERL}$ in Eq.~\ref{eq:erl}, as follows:
\begin{align}
\label{eq:rem}
    \mathcal{L}_{REM} = \mathcal{L}_{MCL} + \lambda \cdot \mathcal{L}_{ERL},
\end{align}
where $\lambda$ is a hyperparameter. We propose Ranked Entropy Minimization, which integrates the advantages of consistency regularization and entropy minimization within a ranked structure based on masking ratios.

\section{Experiments}
In this section, we extensively explore the effectiveness of our REM on CTTA protocol~\cite{Wangetal2022cotta}. The analysis includes comparisons with state-of-the-art baselines, verification of its intended functionality through visualizations, and an understanding of its working mechanisms through ablation studies. Additionally, experiments on Online TTA and Vision-Language Model are provided in~\cref{appendix:tta},~\ref{appendix:tta-clip}.

\subsection{Experimental Setup}
\noindent\textbf{Benchmarks.}
We construct experiments on ImageNet-to-ImageNetC, CIFAR10-to-CIFAR10C, and CIFAR100-to-CIFAR100C. The source domains are ImageNet~\cite{imagenet} and CIFAR~\cite{krizhevsky2009cifar}, while the corresponding robustness benchmarks~\cite{hendrycks2019benchmarking}, ImageNetC, CIFAR10C, and CIFAR100C, are used as the target domains. The suffix C in these datasets indicates corruption, which includes 15 types of corruptions, each with 5 levels of severity. Following~\cite{Wangetal2022cotta,liu2023vida,liu2024continual}, we adopt target domains with level 5 severity across all 15 corruption types for sequential domains. We evaluate the classification error rate for each target domain after adaptation and prediction on target domain data streams in an online manner.

\noindent\textbf{Comparison Methods.}
We compare various types of state-of-the-art CTTA approaches using the ViT-B/16~\cite{dosovitskiy2021vit} pre-trained on the source domain. These include single model-based methods such as Pseudo-label~\cite{Leeetal2013}, Tent~\cite{DequanWangetal2021}, VDP~\cite{gan2023decorate}, and SAR~\cite{niu2023sar}, as well as teacher-student frameworks, including CoTTA~\cite{Wangetal2022cotta}, PETAL~\cite{brahma2023petal}, ViDA~\cite{liu2023vida}, and Continual-MAE~\cite{liu2024continual}. Additionally, the supervised results, obtained by training with target labels using cross-entropy loss, are presented as an upper bound since target labels are unavailable in TTA.

\noindent\textbf{Implementation.}
We implemented the experiments on CIFAR10C and CIFAR100C using the open-source Continual-MAE code and the provided source model weights. For ImageNetC, the experiments were conducted using the open-source ViDA code with ImageNet pre-trained weights from timm~\cite{rw2019timm}. Details for reproducibility and training regimes are provided in~\cref{implementation_details}.

\subsection{Quantitative Results}

\begin{table}[t]
    \vspace{-0.35cm}
    \centering
        \setlength\tabcolsep{0.05cm}
        
        \caption{Forward transfer analysis on ImageNetC. Results (\%) represent the error rates for unseen and seen domains, harmonic mean.}
        \label{tab:dg}
        \begin{adjustbox}
        {width=1\linewidth,center=\linewidth}
        \begin{tabular}{l|ccccc|c|c|c}
        \toprule
         \multirow{2.3}{*}{Method} &  \multicolumn{5}{c|}{\textbf{Directly test on unseen domains}}& \multicolumn{1}{c|}{\textbf{Unseen}}& \multicolumn{1}{c|}{\textbf{Seen}}& \multicolumn{1}{c}{\textbf{Harmonic}} \\ 
         \cmidrule(lr){2-6} \cmidrule(lr){7-9}
         & bri. & contrast & elastic & pixelate & jpeg 
        & Mean$\downarrow$ & Mean$\downarrow$& Mean$\downarrow$\\
        \hline
        Source&26.4&91.4&57.5&38.0&36.2&49.9&58.8&54.0\\
        Tent &25.8&91.9&57.0&37.2&35.7&49.5&54.4&51.8\\
       CoTTA &25.3&88.1&55.7&36.4&34.6&48.0&57.8&52.4\\
       ViDA & 24.6& 68.2& 49.8& \textbf{34.7}& 34.1& 42.3&45.9&44.0\\ 
        \rowcolor{Light!70}REM (lr=1e-4)&\textbf{23.9}&\textbf{66.3}&\textbf{47.6}&35.9&\textbf{33.1}&\textbf{41.4}&45.4&\textbf{43.1}\\
        \rowcolor{Light!70}REM (lr=1e-3)&24.8&66.9&53.5&40.0&39.4&44.9&\textbf{42.1}&43.5\\
        \midrule
        \rowcolor{gray!10}Supervised&22.5&71.1&55.3&38.2&36.6&44.7&38.2&41.2\\
        \hline
        \end{tabular}
        \end{adjustbox}
\end{table}

\noindent\textbf{ImageNet-to-ImageNetC.}~\cref{tab:imagenet} presents the CTTA experimental results for a source model pre-trained on ImageNet, using each corruption in ImageNetC as the target domain. The model sequentially adapts to the target domains over time, and we compare the average error for each domain. Our method improves average performance by 16.6\% over the source model and surpasses the previous state-of-the-art Continual-MAE by 3.3\%. Notably, the performance gap to the supervised learning upper bound is only 3.5\%, demonstrating the effectiveness of our approach.

\noindent\textbf{CIFAR10-to-CIFAR10C and CIFAR100-to-CIFAR100C.} 
\cref{tab:cifar10} and \ref{tab:cifar100} summarize the experimental results for models trained on CIFAR10 and CIFAR100 as source domains, with CIFAR10C and CIFAR100C serving as the respective target domains. The results reveal consistent performance improvements on CIFAR10C and CIFAR100C, which are widely recognized benchmarks for CTTA alongside ImageNetC. These findings highlight the robustness and adaptability of our method across diverse datasets.

\begin{figure}[t]
\begin{center}
\includegraphics[width=0.85\linewidth]{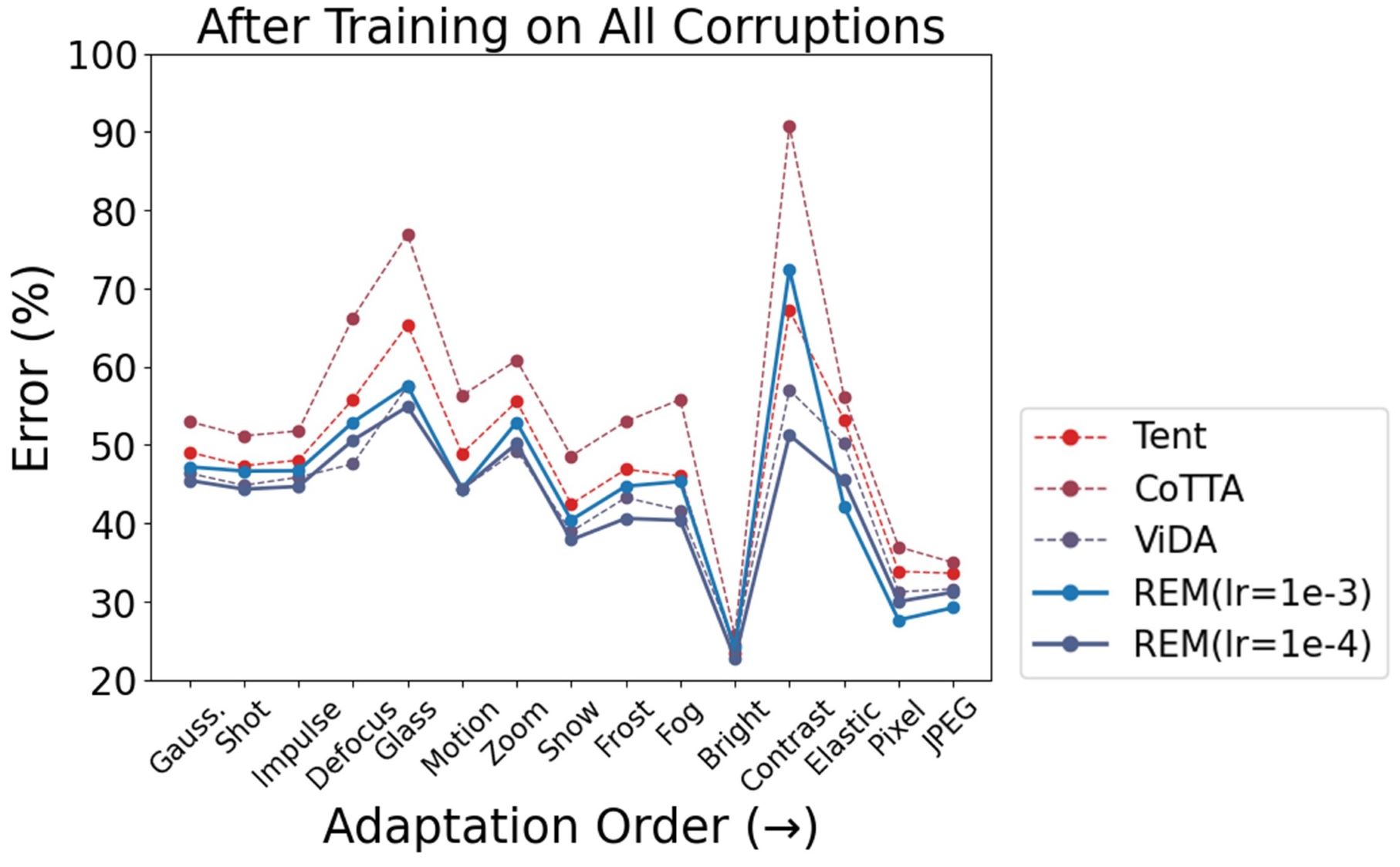}
\end{center}
\caption{Backward transfer analysis on ImageNetC. We compare the performance of CTTA approaches on previous domains.}
\label{fig:forgetting}
\end{figure}

\begin{figure}[t]
\begin{center}
\begin{minipage}{0.48\linewidth}
    \centering
    \includegraphics[width=\linewidth]{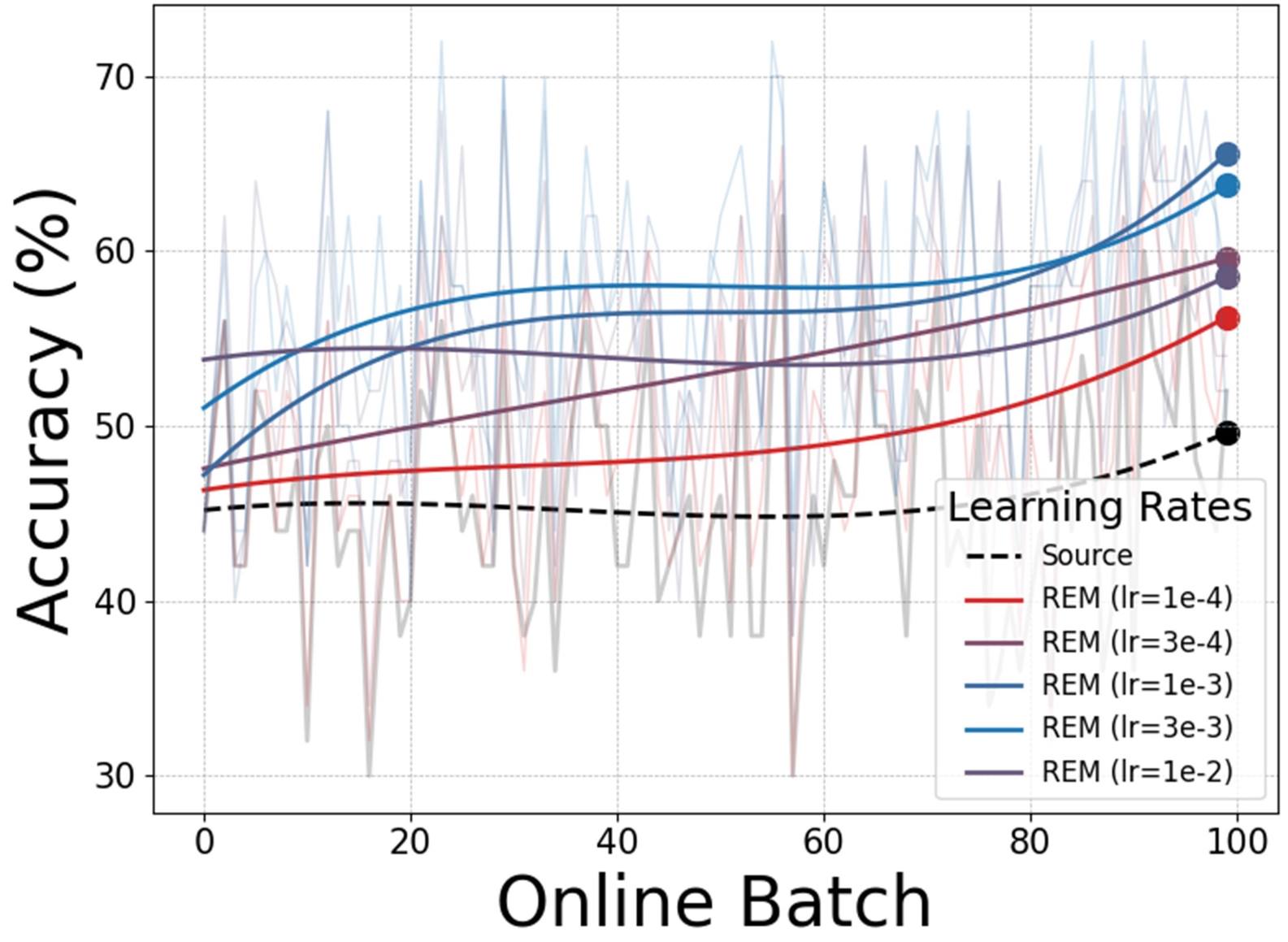}

        \caption{Adaptability analysis on ImageNetC under Gaussian noise corruption.}
    \label{fig:adaptability}
\end{minipage}
\hfill
\begin{minipage}{0.48\linewidth}
    \centering
    \includegraphics[width=\linewidth]{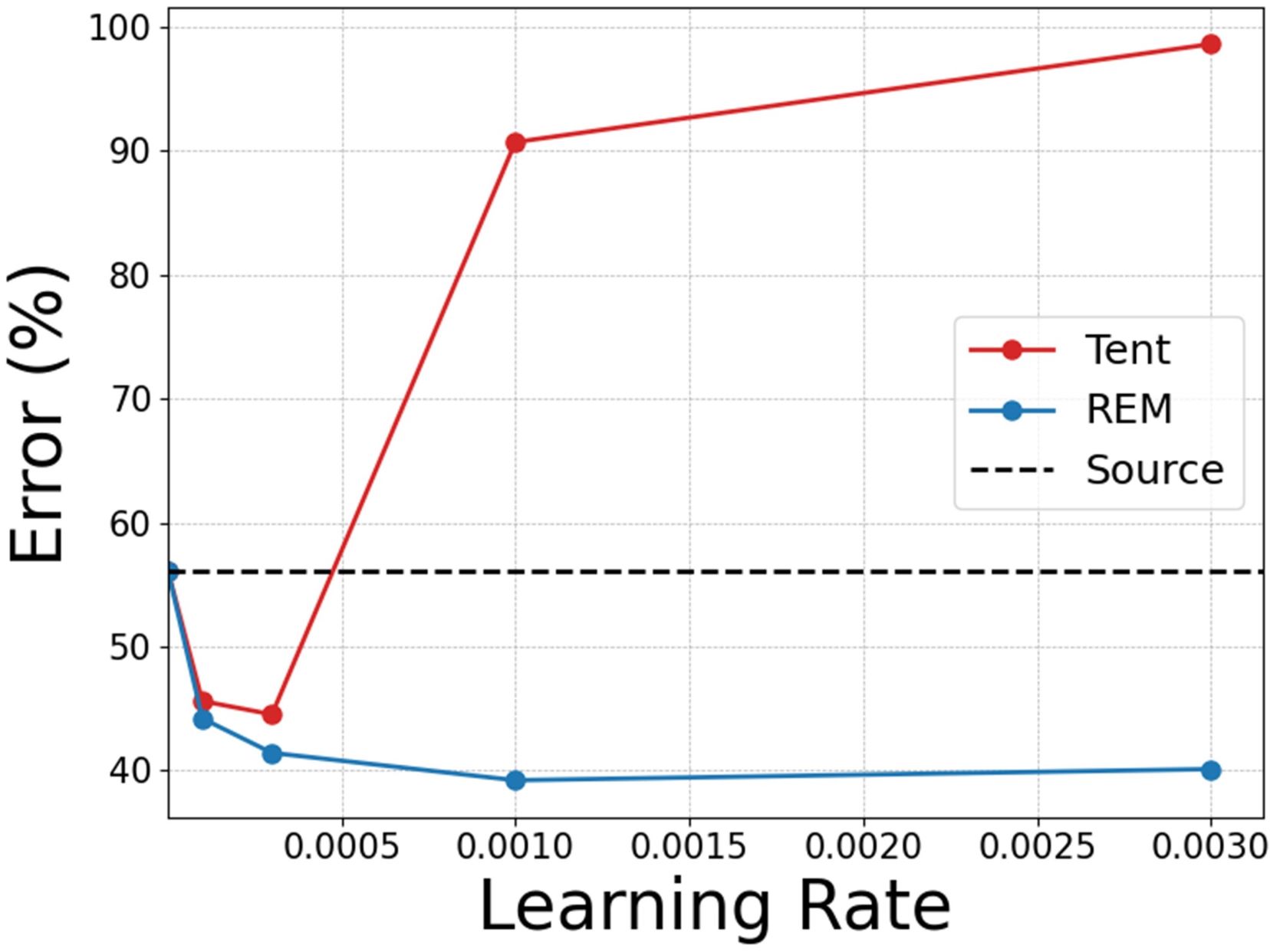}
    \caption{Robustness analysis with respect to learning rate on ImageNetC.}
    \label{fig:learningrate}
\end{minipage}
\end{center}
\end{figure}

\begin{figure*}[t]
\begin{center}
\includegraphics[width=0.98\linewidth]{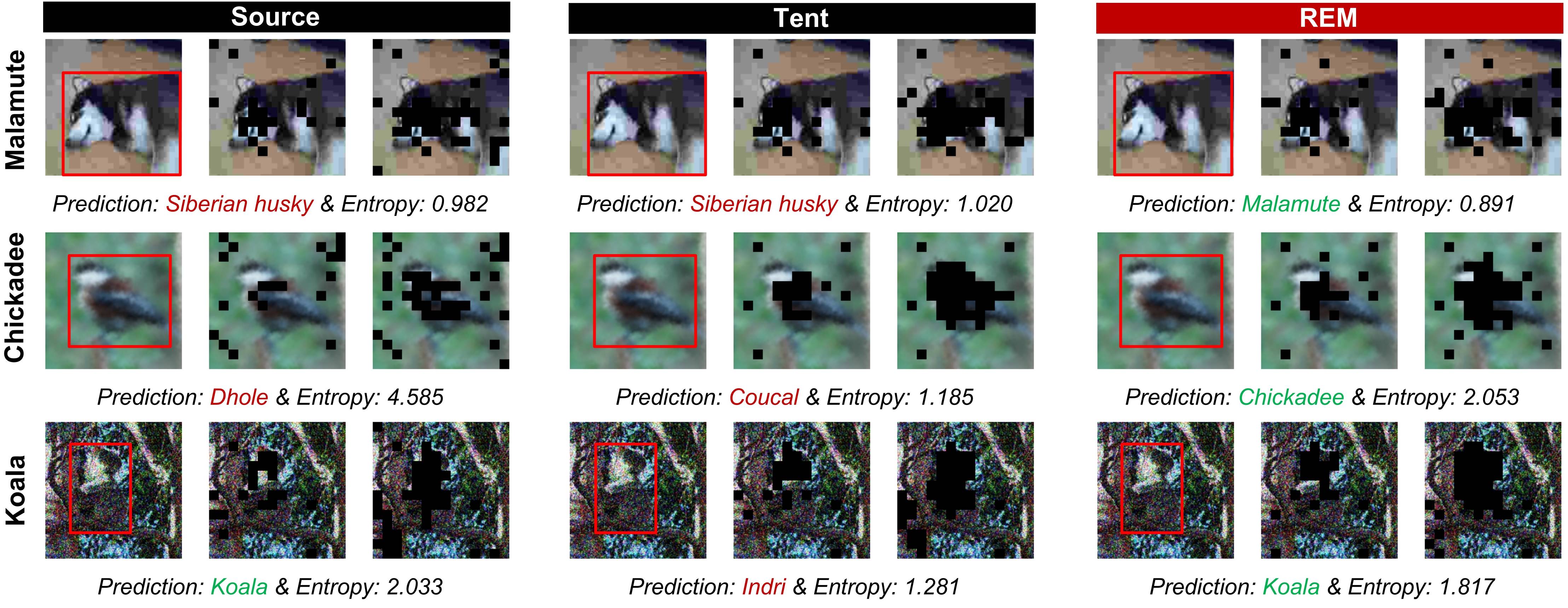}
\end{center}
\caption{\textbf{Masked image visualization.} We compare the predictions and entropy of REM, Tent, and Source and visualize the results of our masking strategy. Each column represents images with masking ratios of 0, 10\%, and 20\% for each method, while each row shows the true label on the left, along with \textcolor{customgreen}{correct} and \textcolor{red!70!black}{incorrect} predictions and its corresponding entropy values.}
\label{fig:mask_visualization}
\end{figure*}

\begin{figure}[t]
\begin{center}
\includegraphics[width=0.93\linewidth]{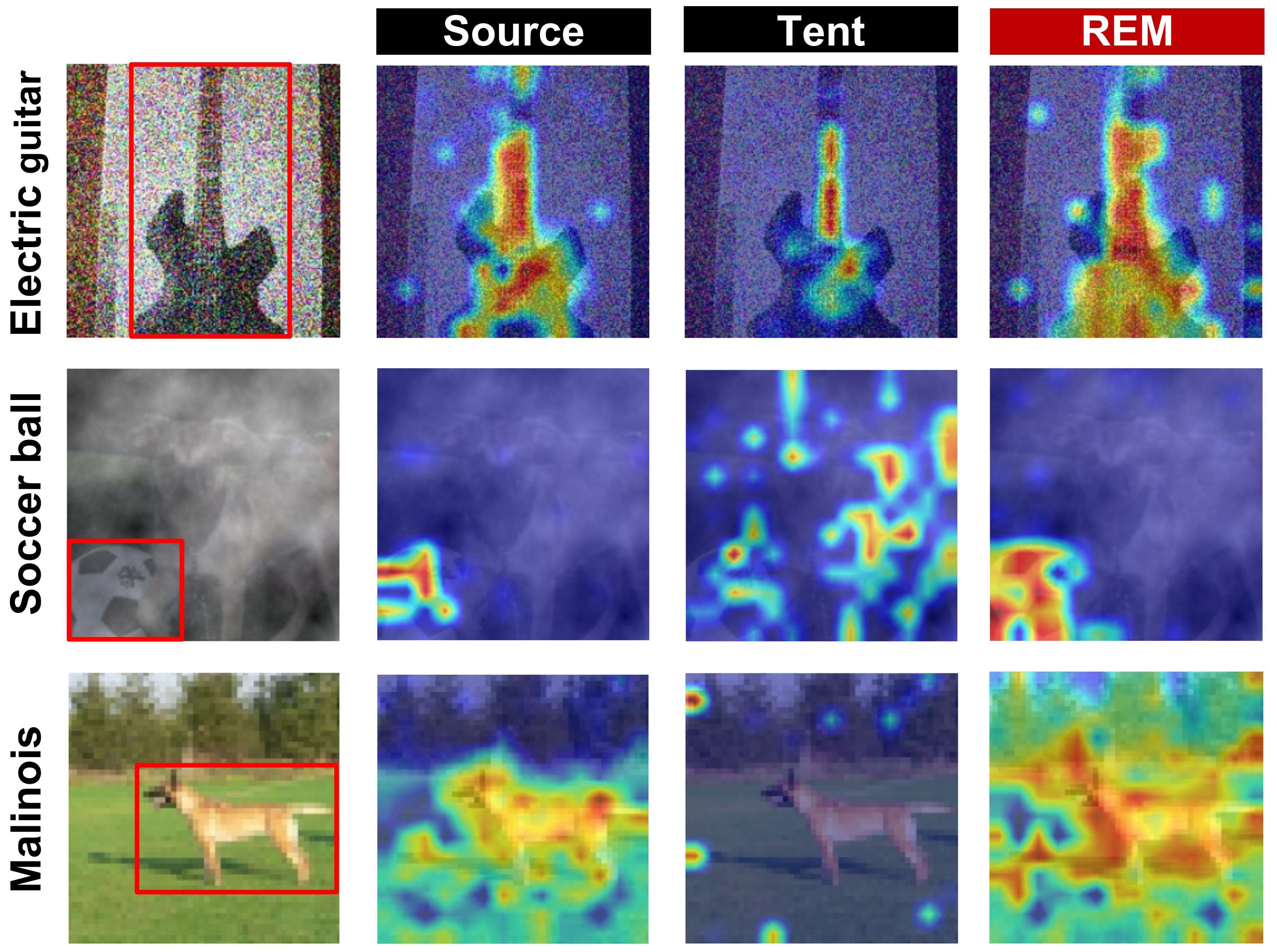}
\end{center}
\caption{\textbf{Grad CAM visualization.} We compare attention maps to identify the pixels contributing to predictions.}
\label{fig:cam_visualization}
\end{figure}

\noindent\textbf{Forward and Backward Transfer Analysis.}
We analyze the performance on both future and past domains during the CTTA process to investigate the potential temporal effects of domain adaptation.~\cref{tab:dg} presents the performance on the 5 unseen domains after training on 10 domains. The investigation of the impact of test-time adaptation at the current time on future performance reveals a trade-off between adaptability and generalization. Rapid adaptation driven by high learning rates achieves the best performance of 42.1\% on Seen domains, while slower adaptation yields the best performance of 41.4\% on Unseen domains. Such a trend is also observed in~\cref{fig:forgetting}, which presents the performance across all domains for the model trained on sequential all domains. Based on the adaptation order, rapid adaptation demonstrates low error on domains learned later, while slow adaptation achieves low error on initial domains.

\noindent\textbf{Discussion.} The preceding experiments indicate that achieving generalized performance across diverse domains does not necessarily guarantee optimal performance in CTTA. Distinctive advantage of TTA over domain generalization lies in its capability to perform domain-specific adaptation through online learning, which is critical for addressing domain shifts effectively.~\cref{fig:adaptability} presents a comparative analysis of adaptability under varying learning rates, while ~\cref{fig:learningrate} illustrates performance trends across a broad range of learning rate settings. The robustness of the proposed method across diverse learning rate boundaries underscores its practical utility, as it enables flexible adaptation speed selection tailored to specific application requirements.

\begin{table}[t]
\centering
\caption{\textbf{Efficiency comparison.} We provide the number of training parameters, total time, number of the forward passes (FP) and models, and error rate (\%).
}
\label{tab:Ablation efficiency}
\resizebox{0.48\textwidth}{!}{ 
\begin{tabular}{l|c|c|c|c|c}
\toprule
Method & Parameters$\downarrow$ & Total Time$\downarrow$ &\# of FP & Total Models & Error \\
\midrule
Source           & 0      & -  & 1  & $M_{src}$       & 55.8 \\
Tent             & 0.04M  & 8m35s  & 1  & $M_{test}$         & 51.0 \\
CoTTA            & 86.4M  & 33m23s  & 3or35 & $M_{test}+M_{ema}+M_{src}$& 54.8 \\
ViDA             & 93.7M  & 54m48s  & 12 & $M_{test}+M_{ema}$ & 43.4 \\
Continual-MAE    & 86.5M  & 59m56s  & 12 & $M_{test}+M_{src}$ & 42.5 \\
\rowcolor{Light!70}REM & 0.03M & 17m21s & 3 & $M_{test}$ & 39.2 \\
\bottomrule
\end{tabular}
}
\end{table}

\begin{figure*}[t]
\begin{center}
\includegraphics[width=0.83\linewidth]{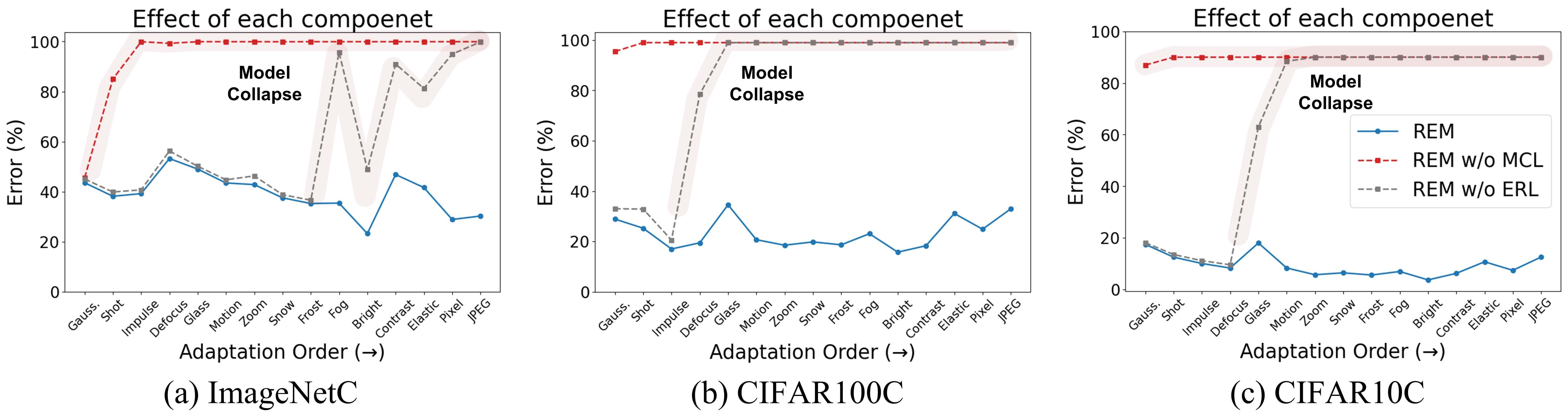}
\end{center}
\vspace{-2mm}
\caption{\textbf{Effect of each component.} We present the results of REM and compare them with variations where MCL and ERL are removed.}
\label{fig:component}
\end{figure*}

\begin{figure*}[t]
\begin{center}
\includegraphics[width=0.83\linewidth]{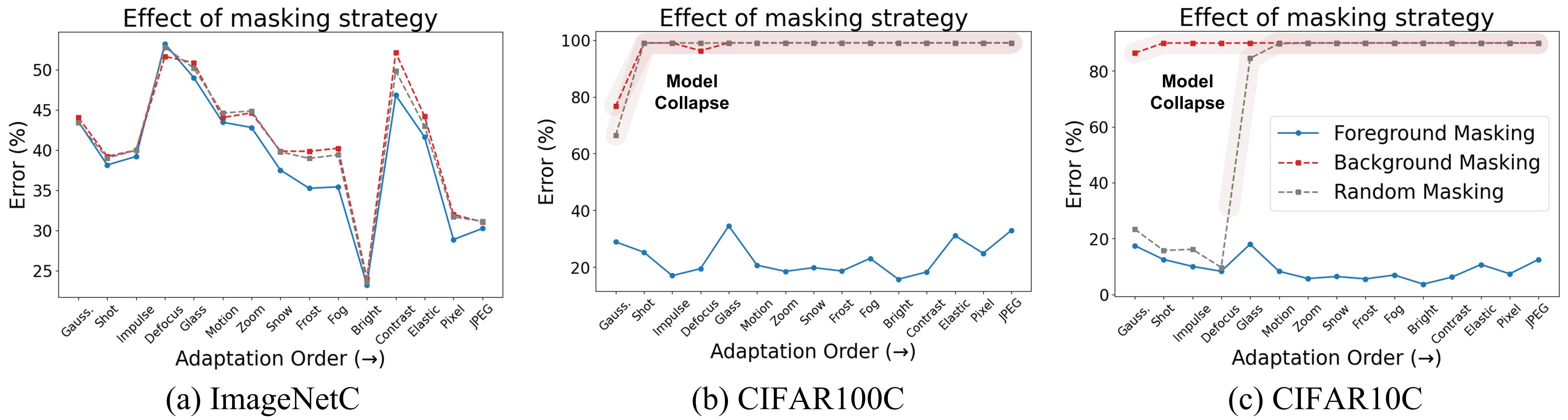}
\end{center}
\vspace{-2mm}
\caption{\textbf{Effect of masking strategy.} We present the results of REM (foreground masking) in comparison with different masking strategies, including background masking, and random masking.}
\label{fig:masking strategy}
\end{figure*}

\subsection{Qualitative Results}
\noindent\textbf{Masked Image Visualization.}
From~\cref{fig:mask_visualization}, we provide a visualization of the masked images generated by the explicit mask chaining, along with the predictions and entropy values for the corresponding original images. It can be observed that masking is specifically applied to the pixels where objects are located. Moreover, compared to Tent, which tends to be overly confident in uncertain predictions, this approach maintains higher entropy, allowing for improvements in the incorrect predictions of the initial source model and effectively mitigating overconfidence.

\noindent\textbf{Class Attention Map Visualization.}
In~\cref{fig:cam_visualization}, we present the Grad-CAM~\cite{selvaraju2017grad} visualization results to highlight the salient pixels influencing the model’s predictions and to gain insights into its decision-making process. Each row corresponds to the results for Gaussian noise, fog, and pixelation, respectively, demonstrating that Tent gradually highlights local regions over time. This phenomenon can be interpreted as the model relying on non-discriminative local features rather than global semantic context for predictions, resulting in the generation of uniform predictions irrespective of the input. In contrast, our method adopts consistency regularization for masked objects and effectively captures comprehensive information related to context.

\subsection{Efficiency Comparison}
\label{Efficiency Comparison}
We compare the computational efficiency of our method with Tent, an EM approach, and CR based state-of-the-art methods such as CoTTA, ViDA, and Continual-MAE in~\cref{tab:Ablation efficiency}. Our method follows the strategy of existing EM approaches that update only the normalization layers of a single test model ($M_{test}$), providing advantages in terms of training time, the number of trainable parameters, and the number of models that need to be stored. Recent CTTA approaches store the EMA model ($M_{ema}$) used as a teacher model and the source model ($M_{src}$) to regress to the initial source weights. In addition, it often requires numerous forward passes to model uncertainty.
Compared to the recent state-of-the-art, Continual-MAE, we achieve a 3.3\% performance improvement while requiring only 30\% of the computation time and 0.03\% of the training parameters.

\subsection{Ablation Studies}
\noindent\textbf{Effect of Each Component.}
We present the results of the ablation experiments for each component in~\cref{fig:component}. Model collapse appeared early for CIFARC, which contains lower resolution and information compared to ImageNetC, when MCL and REL were removed. Note that our method achieves stable performance without model collapse when both are applied, due to the organic design of each method.

\noindent\textbf{Effect of Masking Strategy.}
In order to validate the appropriateness of our masking strategy, \cref{fig:masking strategy} illustrates a comparison between our foreground masking strategy and cases involving background masking or random masking. Our intuition behind defining an explicit ranking relationship for the predictions is satisfied when masking the foreground. In this case, the method performed as designed, and there is no model collapse for all test datasets.

\section{Limitation}
Our study is grounded in the assumption that explicitly masking objects used as the basis for predictions can lead to a decrease in accuracy and an increase in entropy. While our method is simple and intuitive, it is not yet fully supported by rigorous theoretical proof. Despite efforts to address this, the counterexamples arising from the diversity of images still pose significant challenges. To mitigate this, we conduct several experiments that provide empirical evidence demonstrating statistical significance.~\nocite{cai2024sample}

\section{Conclusion}
In this paper, we introduce Ranked Entropy Minimization (REM) to improve stability and efficiency in CTTA. Based on observation of model collapse, we propose a progressive masking strategy and dual complementary loss functions: masked consistency loss and ranked entropy loss. Consequently, REM captures the best of both worlds by integrating the stability of consistency regularization and the efficiency of entropy minimization. Through quantitative evaluations on various CTTA benchmarks, REM achieves state-of-the-art performance, demonstrating its effectiveness. Moreover, extensive qualitative experiments and ablation studies offer in-depth insights into the working principles.
We hope that our work serves as a foundation for valuable discussions on computational cost in CTTA, paving the way for advances in efficiency and real-world applicability.

\section*{Impact Statement}
This paper presents work whose goal is to advance the field
of Machine Learning. There are many potential societal
consequences of our work, none which we feel must be
specifically highlighted here.

\section*{Acknowledgements}
This work was supported by IITP grant funded by the Korea government(MSIT) (RS-2025-02283048, Developing the Next-Generation General AI with Reliability, Ethics, and Adaptability; RS-2023-00236245, Development of Perception/Planning AI SW for Seamless Autonomous Driving in Adverse Weather/Unstructured Environment; RS-2021-II212068, AI Innovation Hub) and NRF-2022R1A2C1091402.

\bibliography{example_paper}

\begin{thebibliography}{51}
\providecommand{\natexlab}[1]{#1}
\providecommand{\url}[1]{\texttt{#1}}
\expandafter\ifx\csname urlstyle\endcsname\relax
  \providecommand{\doi}[1]{doi: #1}\else
  \providecommand{\doi}{doi: \begingroup \urlstyle{rm}\Url}\fi

\bibitem[Alfarra et~al.(2024)Alfarra, Itani, Pardo, shyma~yaser alhuwaider, Ramazanova, Perez, zhipeng cai, M{\"u}ller, and Ghanem]{alfarra2024constraints}
Alfarra, M., Itani, H., Pardo, A., shyma~yaser alhuwaider, Ramazanova, M., Perez, J.~C., zhipeng cai, M{\"u}ller, M., and Ghanem, B.
\newblock Evaluation of test-time adaptation under computational time constraints.
\newblock In \emph{International Conference on Machine Learning}, 2024.

\bibitem[Bolya et~al.(2023)Bolya, Fu, Dai, Zhang, Feichtenhofer, and Hoffman]{bolya2023token}
Bolya, D., Fu, C.-Y., Dai, X., Zhang, P., Feichtenhofer, C., and Hoffman, J.
\newblock Token merging: Your vit but faster.
\newblock In \emph{International Conference on Learning Representations}, 2023.

\bibitem[Boudiaf et~al.(2022)Boudiaf, Mueller, Ben~Ayed, and Bertinetto]{boudiaf2022lame}
Boudiaf, M., Mueller, R., Ben~Ayed, I., and Bertinetto, L.
\newblock Parameter-free online test-time adaptation.
\newblock In \emph{Proceedings of the IEEE/CVF Conference on Computer Vision and Pattern Recognition}, 2022.

\bibitem[Brahma \& Rai(2023)Brahma and Rai]{brahma2023petal}
Brahma, D. and Rai, P.
\newblock A probabilistic framework for lifelong test-time adaptation.
\newblock In \emph{Proceedings of the IEEE/CVF Conference on Computer Vision and Pattern Recognition}, 2023.

\bibitem[Cai et~al.(2024)Cai, Ye, Feng, Qi, and Liu]{cai2024sample}
Cai, C., Ye, Z., Feng, L., Qi, J., and Liu, F.
\newblock Sample-specific masks for visual reprogramming-based prompting.
\newblock In \emph{International Conference on Machine Learning}, 2024.

\bibitem[Csaba et~al.(2024)Csaba, Zhang, M{\"u}ller, Lim, Torr, and Bibi]{csaba2024labeldelay}
Csaba, B., Zhang, W., M{\"u}ller, M., Lim, S.-N., Torr, P., and Bibi, A.
\newblock Label delay in online continual learning.
\newblock In \emph{Advances in Neural Information Processing Systems}, 2024.

\bibitem[Deng et~al.(2009)Deng, Dong, Socher, Li, Li, and Fei-Fei]{imagenet}
Deng, J., Dong, W., Socher, R., Li, L.-J., Li, K., and Fei-Fei, L.
\newblock Imagenet: A large-scale hierarchical image database.
\newblock In \emph{Proceedings of the IEEE/CVF Conference on Computer Vision and Pattern Recognition}, 2009.

\bibitem[D{\"o}bler et~al.(2023)D{\"o}bler, Marsden, and Yang]{dobler2023rmt}
D{\"o}bler, M., Marsden, R.~A., and Yang, B.
\newblock Robust mean teacher for continual and gradual test-time adaptation.
\newblock In \emph{Proceedings of the IEEE/CVF Conference on Computer Vision and Pattern Recognition}, 2023.

\bibitem[Dosovitskiy et~al.(2021)Dosovitskiy, Beyer, Kolesnikov, Weissenborn, Zhai, Unterthiner, Dehghani, Minderer, Heigold, Gelly, Uszkoreit, and Houlsby]{dosovitskiy2021vit}
Dosovitskiy, A., Beyer, L., Kolesnikov, A., Weissenborn, D., Zhai, X., Unterthiner, T., Dehghani, M., Minderer, M., Heigold, G., Gelly, S., Uszkoreit, J., and Houlsby, N.
\newblock An image is worth 16x16 words: Transformers for image recognition at scale.
\newblock In \emph{International Conference on Learning Representations}, 2021.

\bibitem[Fang et~al.(2013)Fang, Xu, and Rockmore]{fang2013vlcs}
Fang, C., Xu, Y., and Rockmore, D.~N.
\newblock Unbiased metric learning: On the utilization of multiple datasets and web images for softening bias.
\newblock In \emph{Proceedings of the IEEE International Conference on Computer Vision}, 2013.

\bibitem[Gan et~al.(2023)Gan, Bai, Lou, Ma, Zhang, Shi, and Luo]{gan2023decorate}
Gan, Y., Bai, Y., Lou, Y., Ma, X., Zhang, R., Shi, N., and Luo, L.
\newblock Decorate the newcomers: Visual domain prompt for continual test time adaptation.
\newblock In \emph{Proceedings of the AAAI Conference on Artificial Intelligence}, volume~37, pp.\  7595--7603, 2023.

\bibitem[Gong et~al.(2022)Gong, Jeong, Kim, Kim, Shin, and Lee]{gong2022note}
Gong, T., Jeong, J., Kim, T., Kim, Y., Shin, J., and Lee, S.-J.
\newblock Note: Robust continual test-time adaptation against temporal correlation.
\newblock \emph{Advances in Neural Information Processing Systems}, 2022.

\bibitem[Goodfellow et~al.(2014)Goodfellow, Pouget-Abadie, Mirza, Xu, Warde-Farley, Ozair, Courville, and Bengio]{goodfellow2014gan}
Goodfellow, I., Pouget-Abadie, J., Mirza, M., Xu, B., Warde-Farley, D., Ozair, S., Courville, A., and Bengio, Y.
\newblock Generative adversarial nets.
\newblock \emph{Advances in neural information processing systems}, 2014.

\bibitem[Hakim et~al.(2024)Hakim, Osowiechi, Noori, Cheraghalikhani, Bahri, Yazdanpanah, Ayed, and Desrosiers]{hakim2024clipartt}
Hakim, G. A.~V., Osowiechi, D., Noori, M., Cheraghalikhani, M., Bahri, A., Yazdanpanah, M., Ayed, I.~B., and Desrosiers, C.
\newblock Clipartt: Light-weight adaptation of clip to new domains at test time.
\newblock \emph{arXiv preprint arXiv:2405.00754}, 2024.

\bibitem[Hendrycks \& Dietterich(2018)Hendrycks and Dietterich]{hendrycks2019benchmarking}
Hendrycks, D. and Dietterich, T.
\newblock Benchmarking neural network robustness to common corruptions and perturbations.
\newblock In \emph{International Conference on Learning Representations}, 2018.

\bibitem[Hendrycks et~al.(2021)Hendrycks, Basart, Mu, Kadavath, Wang, Dorundo, Desai, Zhu, Parajuli, Guo, Song, Steinhardt, and Gilmer]{hendrycks2021inr}
Hendrycks, D., Basart, S., Mu, N., Kadavath, S., Wang, F., Dorundo, E., Desai, R., Zhu, T., Parajuli, S., Guo, M., Song, D., Steinhardt, J., and Gilmer, J.
\newblock The many faces of robustness: A critical analysis of out-of-distribution generalization.
\newblock \emph{ICCV}, 2021.

\bibitem[Hua et~al.(2021)Hua, Wang, Xue, Ren, Wang, and Zhao]{hua2021sslcollapse}
Hua, T., Wang, W., Xue, Z., Ren, S., Wang, Y., and Zhao, H.
\newblock On feature decorrelation in self-supervised learning.
\newblock In \emph{Proceedings of the IEEE/CVF International Conference on Computer Vision}, 2021.

\bibitem[Huggett(2002)]{huggett2002zeno}
Huggett, N.
\newblock Zeno's paradoxes.
\newblock In Zalta, E.~N. (ed.), \emph{The Stanford Encyclopedia of Philosophy}. Metaphysics Research Lab, Stanford University, 2018 edition, 2002.
\newblock \url{https://plato.stanford.edu/ENTRIES/paradox-zeno/}.

\bibitem[Iwasawa \& Matsuo(2021)Iwasawa and Matsuo]{iwasawa2021t3a}
Iwasawa, Y. and Matsuo, Y.
\newblock Test-time classifier adjustment module for model-agnostic domain generalization.
\newblock \emph{Advances in Neural Information Processing Systems}, 2021.

\bibitem[Kar et~al.(2022)Kar, Yeo, Atanov, and Zamir]{kar20223dcc}
Kar, O.~F., Yeo, T., Atanov, A., and Zamir, A.
\newblock 3d common corruptions and data augmentation.
\newblock In \emph{Proceedings of the IEEE/CVF Conference on Computer Vision and Pattern Recognition}, 2022.

\bibitem[Krizhevsky et~al.(2009)Krizhevsky, Hinton, et~al.]{krizhevsky2009cifar}
Krizhevsky, A., Hinton, G., et~al.
\newblock Learning multiple layers of features from tiny images.
\newblock 2009.

\bibitem[Lee(2013)]{Leeetal2013}
Lee, D.-H.
\newblock Pseudo-label : The simple and efficient semi-supervised learning method for deep neural networks.
\newblock In \emph{International conference on machine learning Workshop}, 2013.

\bibitem[Lee et~al.(2024)Lee, Jung, Lee, Park, Shin, Hwang, and Yoon]{lee2024deyo}
Lee, J., Jung, D., Lee, S., Park, J., Shin, J., Hwang, U., and Yoon, S.
\newblock Entropy is not enough for test-time adaptation: From the perspective of disentangled factors.
\newblock In \emph{International Conference on Learning Representations}, 2024.

\bibitem[Lee \& Chang(2024)Lee and Chang]{lee2024cmf}
Lee, J.-H. and Chang, J.-H.
\newblock Continual momentum filtering on parameter space for online test-time adaptation.
\newblock In \emph{The Twelfth International Conference on Learning Representations}, 2024.

\bibitem[Li et~al.(2017)Li, Yang, Song, and Hospedales]{li2017pacs}
Li, D., Yang, Y., Song, Y.-Z., and Hospedales, T.~M.
\newblock Deeper, broader and artier domain generalization.
\newblock In \emph{Proceedings of the IEEE international conference on computer vision}, 2017.

\bibitem[Liu et~al.(2024{\natexlab{a}})Liu, Xu, Yang, Zhang, Zhang, Chen, Guo, and Zhang]{liu2024continual}
Liu, J., Xu, R., Yang, S., Zhang, R., Zhang, Q., Chen, Z., Guo, Y., and Zhang, S.
\newblock Continual-mae: Adaptive distribution masked autoencoders for continual test-time adaptation.
\newblock \emph{Proceedings of the IEEE/CVF Conference on Computer Vision and Pattern Recognition}, 2024{\natexlab{a}}.

\bibitem[Liu et~al.(2024{\natexlab{b}})Liu, Yang, Jia, Lu, Guo, Xue, and Zhang]{liu2023vida}
Liu, J., Yang, S., Jia, P., Lu, M., Guo, Y., Xue, W., and Zhang, S.
\newblock Vida: Homeostatic visual domain adapter for continual test time adaptation.
\newblock In \emph{International Conference on Learning Representations}, 2024{\natexlab{b}}.

\bibitem[Marsden et~al.(2024)Marsden, D{\"o}bler, and Yang]{marsden2024roid}
Marsden, R.~A., D{\"o}bler, M., and Yang, B.
\newblock Universal test-time adaptation through weight ensembling, diversity weighting, and prior correction.
\newblock In \emph{Proceedings of the IEEE/CVF Winter Conference on Applications of Computer Vision}, 2024.

\bibitem[Moon et~al.(2020)Moon, Kim, Shin, and Hwang]{moon2020confidence}
Moon, J., Kim, J., Shin, Y., and Hwang, S.
\newblock Confidence-aware learning for deep neural networks.
\newblock In \emph{international conference on machine learning}, 2020.

\bibitem[Naeini et~al.(2015)Naeini, Cooper, and Hauskrecht]{naeini2015ece}
Naeini, M.~P., Cooper, G., and Hauskrecht, M.
\newblock Obtaining well calibrated probabilities using bayesian binning.
\newblock In \emph{Proceedings of the AAAI conference on artificial intelligence}, 2015.

\bibitem[Niu et~al.(2022)Niu, Wu, Zhang, Chen, Zheng, Zhao, and Tan]{niu2022eata}
Niu, S., Wu, J., Zhang, Y., Chen, Y., Zheng, S., Zhao, P., and Tan, M.
\newblock Efficient test-time model adaptation without forgetting.
\newblock In \emph{International conference on machine learning}. PMLR, 2022.

\bibitem[Niu et~al.(2023)Niu, Wu, Zhang, Wen, Chen, Zhao, and Tan]{niu2023sar}
Niu, S., Wu, J., Zhang, Y., Wen, Z., Chen, Y., Zhao, P., and Tan, M.
\newblock Towards stable test-time adaptation in dynamic wild world.
\newblock In \emph{Internetional Conference on Learning Representations}, 2023.

\bibitem[Niu et~al.(2024)Niu, Miao, Chen, Wu, and Zhao]{niu2024foa}
Niu, S., Miao, C., Chen, G., Wu, P., and Zhao, P.
\newblock Test-time model adaptation with only forward passes.
\newblock In \emph{International Conference on Machine Learning}, 2024.

\bibitem[Noh et~al.(2023)Noh, Park, Lee, and Ham]{noh2023rankmixup}
Noh, J., Park, H., Lee, J., and Ham, B.
\newblock Rankmixup: Ranking-based mixup training for network calibration.
\newblock In \emph{Proceedings of the IEEE/CVF International Conference on Computer Vision}, 2023.

\bibitem[Osowiechi et~al.(2024)Osowiechi, Noori, Hakim, Yazdanpanah, Bahri, Cheraghalikhani, Dastani, Beizaee, Ayed, and Desrosiers]{osowiechi2024watt}
Osowiechi, D., Noori, M., Hakim, G. A.~V., Yazdanpanah, M., Bahri, A., Cheraghalikhani, M., Dastani, S., Beizaee, F., Ayed, I.~B., and Desrosiers, C.
\newblock {WATT}: Weight average test time adaptation of {CLIP}.
\newblock In \emph{Advances in neural information processing systems}, 2024.

\bibitem[Peng et~al.(2018)Peng, Usman, Kaushik, Wang, Hoffman, and Saenko]{peng2018visda}
Peng, X., Usman, B., Kaushik, N., Wang, D., Hoffman, J., and Saenko, K.
\newblock Visda: A synthetic-to-real benchmark for visual domain adaptation.
\newblock In \emph{Proceedings of the IEEE Conference on Computer Vision and Pattern Recognition Workshops}, pp.\  2021--2026, 2018.

\bibitem[Radford et~al.(2021)Radford, Kim, Hallacy, Ramesh, Goh, Agarwal, Sastry, Askell, Mishkin, Clark, et~al.]{radford2021clip}
Radford, A., Kim, J.~W., Hallacy, C., Ramesh, A., Goh, G., Agarwal, S., Sastry, G., Askell, A., Mishkin, P., Clark, J., et~al.
\newblock Learning transferable visual models from natural language supervision.
\newblock In \emph{International conference on machine learning}, 2021.

\bibitem[Recht et~al.(2019)Recht, Roelofs, Schmidt, and Shankar]{recht2019inv2}
Recht, B., Roelofs, R., Schmidt, L., and Shankar, V.
\newblock Do imagenet classifiers generalize to imagenet?
\newblock In \emph{International conference on machine learning}. PMLR, 2019.

\bibitem[Selvaraju et~al.(2017)Selvaraju, Cogswell, Das, Vedantam, Parikh, and Batra]{selvaraju2017grad}
Selvaraju, R.~R., Cogswell, M., Das, A., Vedantam, R., Parikh, D., and Batra, D.
\newblock Grad-cam: Visual explanations from deep networks via gradient-based localization.
\newblock In \emph{Proceedings of the IEEE international conference on computer vision}, 2017.

\bibitem[Shimodaira(2000)]{domainshift}
Shimodaira, H.
\newblock Improving predictive inference under covariate shift by weighting the log-likelihood function.
\newblock \emph{Journal of statistical planning and inference}, 2000.

\bibitem[Shu et~al.(2022)Shu, Nie, Huang, Yu, Goldstein, Anandkumar, and Xiao]{shu2022tpt}
Shu, M., Nie, W., Huang, D.-A., Yu, Z., Goldstein, T., Anandkumar, A., and Xiao, C.
\newblock Test-time prompt tuning for zero-shot generalization in vision-language models.
\newblock \emph{Advances in Neural Information Processing Systems}, 2022.

\bibitem[Son et~al.(2024)Son, Ryu, Lee, and Lee]{son2023maskedkd}
Son, S., Ryu, J., Lee, N., and Lee, J.
\newblock The role of masking for efficient supervised knowledge distillation of vision transformers.
\newblock In \emph{European Conference on Computer Vision}, 2024.

\bibitem[Sun et~al.(2020)Sun, Wang, Liu, Miller, Efros, and Hardt]{sun2020ttt}
Sun, Y., Wang, X., Liu, Z., Miller, J., Efros, A., and Hardt, M.
\newblock Test-time training with self-supervision for generalization under distribution shifts.
\newblock In \emph{International conference on machine learning}. PMLR, 2020.

\bibitem[Tarvainen \& Valpola(2017)Tarvainen and Valpola]{tarvainen2017meanteacher}
Tarvainen, A. and Valpola, H.
\newblock Mean teachers are better role models: Weight-averaged consistency targets improve semi-supervised deep learning results.
\newblock \emph{Advances in neural information processing systems}, 2017.

\bibitem[Venkateswara et~al.(2017)Venkateswara, Eusebio, Chakraborty, and Panchanathan]{venkateswara2017officehome}
Venkateswara, H., Eusebio, J., Chakraborty, S., and Panchanathan, S.
\newblock Deep hashing network for unsupervised domain adaptation.
\newblock In \emph{Proceedings of the IEEE conference on computer vision and pattern recognition}, 2017.

\bibitem[Wang et~al.(2021)Wang, Shelhamer, Liu, Olshausen, and Darrell]{DequanWangetal2021}
Wang, D., Shelhamer, E., Liu, S., Olshausen, B.~A., and Darrell, T.
\newblock Tent: Fully test-time adaptation by entropy minimization.
\newblock In \emph{International Conference on Learning Representations}, 2021.

\bibitem[Wang et~al.(2019)Wang, Ge, Lipton, and Xing]{wang2019ins}
Wang, H., Ge, S., Lipton, Z., and Xing, E.~P.
\newblock Learning robust global representations by penalizing local predictive power.
\newblock In \emph{Advances in Neural Information Processing Systems}, 2019.

\bibitem[Wang et~al.(2022)Wang, Fink, Van~Gool, and Dai]{Wangetal2022cotta}
Wang, Q., Fink, O., Van~Gool, L., and Dai, D.
\newblock Continual test-time domain adaptation.
\newblock In \emph{Proceedings of Conference on Computer Vision and Pattern Recognition}, 2022.

\bibitem[Wightman(2019)]{rw2019timm}
Wightman, R.
\newblock Pytorch image models.
\newblock \url{https://github.com/rwightman/pytorch-image-models}, 2019.

\bibitem[Zhang et~al.(2022)Zhang, Levine, and Finn]{zhang2022memo}
Zhang, M., Levine, S., and Finn, C.
\newblock Memo: Test time robustness via adaptation and augmentation.
\newblock \emph{Advances in neural information processing systems}, 2022.

\bibitem[Zhang et~al.(2025)Zhang, Bian, Kong, Zhao, and Zhang]{zhang2024come}
Zhang, Q., Bian, Y., Kong, X., Zhao, P., and Zhang, C.
\newblock {COME}: Test-time adaption by conservatively minimizing entropy.
\newblock In \emph{International Conference on Learning Representations}, 2025.

\end{thebibliography}
\bibliographystyle{icml2025}

\newpage
\appendix
\onecolumn

\section{Overall Framework and Comparison with Other Approaches}
\begin{figure*}[t]
\begin{center}
\includegraphics[width=0.8\linewidth]{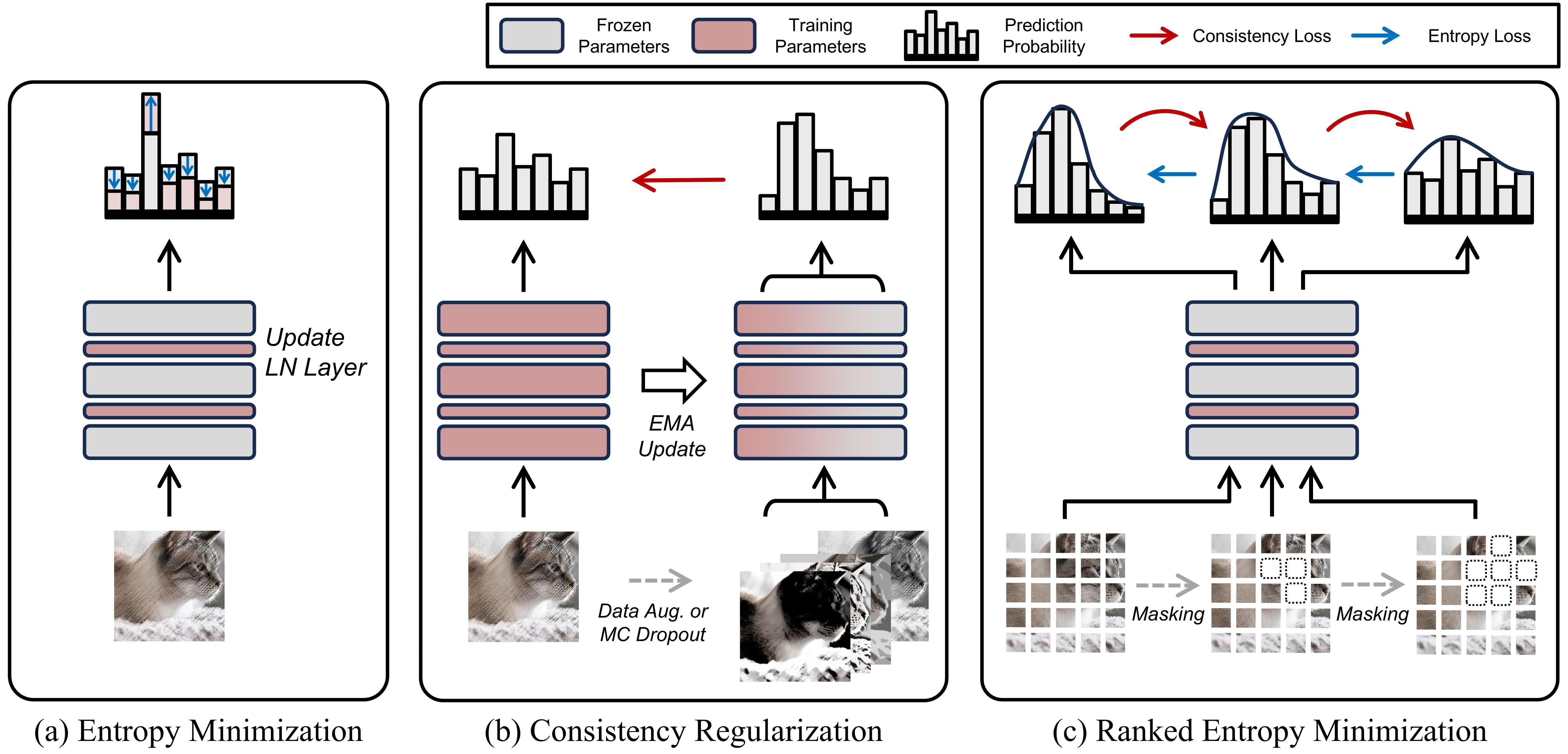}
\end{center}
\caption{\textbf{Conceptual illustration comparing CTTA frameworks.} (a) Entropy minimization approach updates only the normalization layer for the target domain while utilizing a single model.
(b) Consistency regularization method employs both a student model, which updates all parameters, and a teacher model, which is updated via exponential moving average (EMA). This approach enhances prediction diversity by numerous data augmentations or applying MC dropout to the model.
(c) Our proposed method updates the normalization layer and adopts a single model. It integrates entropy minimization and consistency regularization with only two additional forward passes.}
\label{fig:framework}
\end{figure*}

To improve clarity and facilitate comprehension, we present the overall frameworks of the Entropy Minimization (EM) approach, the Consistency Regularization (CR) approach, and our proposed method, REM in~\cref{fig:framework}. We maintain the training scheme of EM to preserve its computational efficiency. Additionally, we replace the traditional data augmentations used in CR approaches with explicit masking, eliminating the reliance on extensive image augmentations for enhancing predictive diversity. Consequently, our method preserves the computational efficiency of EM approaches while enhancing the robustness and generalization performance of CR approaches.

\begin{table}[b]
\centering
\vspace{-2mm}
\caption{Implementation details.
}
\label{tab:implementation_details}
\resizebox{0.95\textwidth}{!}{ 
\begin{tabular}{l|cc|cc|cc}
\toprule
Task & \multicolumn{2}{c|}{CTTA} &\multicolumn{2}{c|}{TTA} & \multicolumn{2}{c}{TTA-CLIP}\\
\midrule
Dataset &ImageNetC&CIFARC&ImageNetC&ImageNet-R/V2/S&CIFAR&Other Datasets\\
\midrule
\rowcolor{Light!60}\multicolumn{7}{c}{\textbf{Experimental Protocols}}\\
Reproducibility& \multicolumn{2}{c|}{Continual-MAE~\cite{liu2024continual}}& \multicolumn{2}{c|}{DeYO~\cite{lee2024deyo}}& \multicolumn{2}{c}{WATT~\cite{osowiechi2024watt}}\\
\rowcolor{Light!60}\multicolumn{7}{c}{\textbf{Training Parameters}}\\
Optimizer              & Adam & Adam  & SGD  & SGD  & Adam  & Adam\\
Optimizer momentum     & (0.9, 0.999) & (0.9, 0.999)  & 0.9  & 0.9  & (0.9, 0.999)  & (0.9, 0.999) \\
Learning rate          & 1e-3 & 1e-3  & 1e-3  & 1e-3  & 1e-3  & 1e-4 \\
Batch size             & 50 & 20  & 1 or 64  & 1 or 64  & 128  & 128 \\
Model architecture             & ViT-B/16 & ViT-B/16  & ViT-B/16  & ViT-B/16 & CLIP-ViT-B/16 & CLIP-ViT-B/16 \\
\rowcolor{Light!60}\multicolumn{7}{c}{\textbf{Algorithm Parameters}}\\
$\lambda$ (Eq.~\ref{eq:rem}) & 1.0 & 1.0 & 0.5 & 0.5 & 1.0 & 1.0 \\
$\mathsf{m}$ (Eq.~\ref{eq:erl}) & 0 & 0 & 0 & 0 & 0 & 0 \\

\bottomrule
\end{tabular}
}
\end{table}

\section{Implementation Details}
\label{implementation_details}
\cref{tab:implementation_details} provides details on the implementation of our experiments, including optimizer settings, learning rates, batch sizes, model architectures, and hyperparameters. For CTTA experiments, we follow the Continual-MAE framework. Specifically, for CIFAR datasets, we resize the input images to 384×384, while for all other experiments, the images are resized to 224×224.

\begin{table*}[t]
\centering
\caption{\label{tab:label_shift}Classification accuracy (\%) on ImageNetC (severity level5) under online imbalanced label shifts (imbalance ratio $= \infty$).}

\small
\setlength\tabcolsep{2pt}
\begin{adjustbox}{width=1\linewidth,center=\linewidth}
\begin{tabular}{l|ccccccccccccccc|cc}
\hline
Label Shifts &
 \rotatebox[origin=c]{50}{Gaussian} & \rotatebox[origin=c]{50}{shot} & \rotatebox[origin=c]{50}{impulse} & \rotatebox[origin=c]{50}{defocus} & \rotatebox[origin=c]{50}{glass} & \rotatebox[origin=c]{50}{motion} & \rotatebox[origin=c]{50}{zoom} & \rotatebox[origin=c]{50}{snow} & \rotatebox[origin=c]{50}{frost} & \rotatebox[origin=c]{50}{fog}  & \rotatebox[origin=c]{50}{brightness} & \rotatebox[origin=c]{50}{contrast} & \rotatebox[origin=c]{50}{elastic\_trans} & \rotatebox[origin=c]{50}{pixelate} & \rotatebox[origin=c]{50}{jpeg}
& Mean$\uparrow$ & Gain\\\hline
Source~\cite{dosovitskiy2021vit}&9.4 &6.7 &8.3 &29.1 &23.4 &34.0 &27.0 &15.8 &26.3 &47.4 &54.7 &43.9 &30.5 &44.5 &47.6 &29.9 &0.0\\
MEMO~\cite{zhang2022memo}&21.6 &17.4 &20.6 &37.1 &29.6 &40.6 &34.4 &25.0 &34.8 &55.2 &65.0 &54.9 &37.4 &55.5 &57.7 &39.1&9.2\\
Tent~\cite{DequanWangetal2021}&53.1&53.1&54.3&54.2&51.5&58.6&52.4&3.5&7.8&69.5&74.8&67.0&58.7&69.2&66.2&52.9&23.0\\
EATA~\cite{niu2022eata}&45.5&47.2&44.1&45.4&41.5&52.0&47.4&54.8&46.7&57.1&70.4&29.2&55.9&62.2&60.6&50.7&20.8\\
SAR~\cite{niu2023sar} &53.1&53.3&54.3&54.0&52.1&58.0&52.7&8.6&28.6&69.1&74.7&66.7&59.1&67.1&64.9&54.4&24.5\\
DeYO~\cite{lee2024deyo} &52.9&54.8&55.4&54.1&55.6&62.1&34.4&64.6&63.7&71.1&77.1&64.2&\textbf{67.2}&72.4&68.2&61.2&31.3\\
\rowcolor{Light!70}REM (Ours)
&\textbf{57.0}	&\textbf{57.2}	&\textbf{58.1}	&\textbf{58.6}	&\textbf{56.3}	&\textbf{63.2}	&\textbf{58.4}	&\textbf{67.3}	&\textbf{67.5}	&\textbf{74.4}	&\textbf{78.9}	&\textbf{70.5}	&65.6 &\textbf{73.3}	&\textbf{70.0}	&\textbf{65.1} &\textbf{35.2}\\

\hline
\end{tabular}
\end{adjustbox}
\end{table*}

\begin{table*}[t]
\centering
\caption{\label{tab:batch_1}Classification accuracy (\%) on ImageNetC (severity level5) under batch size 1.}

\small
\setlength\tabcolsep{2pt}
\begin{adjustbox}{width=1\linewidth,center=\linewidth}
\begin{tabular}{l|ccccccccccccccc|cc}
\hline
Batch Size 1 &
 \rotatebox[origin=c]{50}{Gaussian} & \rotatebox[origin=c]{50}{shot} & \rotatebox[origin=c]{50}{impulse} & \rotatebox[origin=c]{50}{defocus} & \rotatebox[origin=c]{50}{glass} & \rotatebox[origin=c]{50}{motion} & \rotatebox[origin=c]{50}{zoom} & \rotatebox[origin=c]{50}{snow} & \rotatebox[origin=c]{50}{frost} & \rotatebox[origin=c]{50}{fog} & \rotatebox[origin=c]{50}{brightness} & \rotatebox[origin=c]{50}{contrast} & \rotatebox[origin=c]{50}{elastic\_trans} & \rotatebox[origin=c]{50}{pixelate} & \rotatebox[origin=c]{50}{jpeg}
& Mean$\uparrow$ & Gain\\\hline
Source~\cite{dosovitskiy2021vit}&9.5 &6.8 &8.2 &29.0 &23.5 &33.9 &27.1 &15.9 &26.5 &47.2 &54.7 &44.1 &30.5 &44.5 &47.8 &29.9 &0.0\\
MEMO~\cite{zhang2022memo}&21.6 &17.4 &20.6 &37.1 &29.6 &40.6 &34.4 &25.0 &34.8 &55.2 &65.0 &54.9 &37.4 &55.5 &57.7 &39.1&9.2\\
Tent~\cite{DequanWangetal2021}&52.1&51.8&53.2&52.4&48.7&56.5&49.5&8.5&15.2&67.3&73.4&66.7&52.6&64.9&64.3&51.8&21.9\\
EATA~\cite{niu2022eata}&48.5&46.5&49.6&46.2&40.2&50.5&44.1&37.8&41.7&64.6&68.2&64.5&49.6&61.0&61.6&51.6&21.7\\
SAR~\cite{niu2023sar} &52.0&51.7&53.1&51.7&48.9&56.8&50.6&16.8&54.8&67.2&74.7&66.1&55.3&66.8&65.2&55.5&25.6\\
DeYO~\cite{lee2024deyo} &54.6&55.6&56.0&55.5&17.3&62.7&\textbf{59.5}&65.6&64.4&72.0&77.3&10.9&\textbf{66.3}&71.8&68.7&57.2&27.3 \\
\rowcolor{Light!70}REM (Ours)
&\textbf{57.4}	&\textbf{57.8}	&\textbf{58.6}	&\textbf{59.2}	&\textbf{56.9}	&\textbf{63.5}	&59.1	&\textbf{68.4}	&\textbf{67.5}	&\textbf{74.5}	&\textbf{79.0}	&\textbf{71.1}	&65.7	&\textbf{73.3}	&\textbf{70.5}	&\textbf{65.5} &\textbf{35.6}\\
\hline
\end{tabular}
\end{adjustbox}
\end{table*}

\begin{table*}[hbt!]
\begin{center}
    \begin{minipage}{0.47\textwidth}
    \centering
        \caption{\label{tab:mixed_shift}Classification accuracy (\%) on ImageNetC (severity level 5 and level 3) under mixture of 15 corruption.}
        \setlength\tabcolsep{0.043cm}
        \small
        \begin{tabular}{l|cc}
        \hline
        Mixed Shifts &
        Level 5 & Level 3\\\hline
        Source~\cite{dosovitskiy2021vit}&29.9 &53.8\\
        Tent~\cite{DequanWangetal2021}&24.1 &70.2\\
        EATA~\cite{niu2022eata}&56.4 &69.6\\
        SAR~\cite{niu2023sar} &57.1 &70.7\\
        DeYO~\cite{lee2024deyo} &59.4 &72.1\\
        \rowcolor{Light!70}REM (Ours)&\textbf{62.4} &\textbf{74.0}\\
        \hline
        \end{tabular}
        
    \end{minipage}
    \hspace{0.02\textwidth}
    \begin{minipage}{0.47\textwidth}
        \centering
        \caption{\label{tab:domain}Classification accuracy (\%) on ImagNet-R/V2/Sketch. Mean (\%) denotes the average accuracy across 3 target domains.}
        \setlength\tabcolsep{0.05cm}
        \small
        \begin{tabular}{l|ccc|c}
        \hline
        Domain Shifts &
        R &V2 &Sketch &Mean \\\hline
        Source~\cite{dosovitskiy2021vit}&59.5 &\textbf{75.4} &44.9 &59.9\\
        Tent~\cite{DequanWangetal2021}&63.9 &75.2 &49.1 &62.7\\
        CoTTA~\cite{Wangetal2022cotta} &63.5 &\textbf{75.4} &\textbf{50.0} &62.9\\
        
        SAR~\cite{niu2023sar} &63.3 &75.1 &48.7 &62.4\\
        FOA~\cite{niu2024foa} &63.8 &\textbf{75.4} &49.9 &63.0\\
        \rowcolor{Light!70}REM (Ours)&\textbf{64.3}&75.2&49.7 &\textbf{63.1}\\
        \hline
        \end{tabular}
    \end{minipage}
\end{center}
\end{table*}

\section{Experiments on Online Test-Time Adaptation Scenario}
\label{appendix:tta}
In addition to CTTA scenarios, our method is readily applicable to a wide range of TTA scenarios. To evaluate its effectiveness in a more challenging setting, we compare our approach against EM-based state-of-the-art methods, including MEMO~\cite{zhang2022memo}, Tent~\cite{DequanWangetal2021}, EATA~\cite{niu2022eata}, SAR~\cite{niu2023sar}, and DeYO~\cite{lee2024deyo}, in the wild online TTA scenarios proposed in SAR.

\textbf{Online Imbalanced Label Distribution Shifts.}
~\cref{tab:label_shift} presents the performance comparison of TTA methods under class-imbalanced distributions across different domains. Our method achieves best performance across all domains except for elastic transform, improving the average performance by 3.9\% compared to the previous state-of-the-art method, DeYO.

\textbf{Batch Size 1.}~\cref{tab:batch_1} shows the results for TTA under a batch size of 1, demonstrating the robustness of our method in scenarios where batch statistics cannot be effectively leveraged. Similar to the label shift scenario, our method achieves the best performance across all domains except for elastic transform, resulting in an 8.3\% performance improvement.

\textbf{Mixed Distribution Shifts.}~\cref{tab:mixed_shift} presents the results for the TTA scenario where domain boundaries are ambiguous, leading to mixed domain distributions. Our method achieves performance improvements of 3.0\% and 1.9\% for corruption severity levels 5 and 3, respectively, indicating its potential for generalization across diverse domains.

\textbf{Domain Shifts.}~\cref{tab:domain} presents the TTA results for domain shifts from ImageNet to ImageNet-R~\cite{hendrycks2021inr}, ImageNet-V2~\cite{recht2019inv2}, and ImageNet-Sketch~\cite{wang2019ins}. We achieve a mean accuracy of 63.1\% across all domains, surpassing the previous state-of-the-art FOA~\cite{niu2024foa}, which achieved 63.0\%.

\section{Experiments on Vision-Language Model}
\label{appendix:tta-clip}
Our proposed REM can be applied in a plug-and-play manner and is adaptable to various modalities. We present experiments on the TTA setting with CLIP as the target model in~\cref{tab:vlm}. We compare the performance of our method against CLIP~\cite{radford2021clip}, Tent~\cite{DequanWangetal2021}, TPT~\cite{shu2022tpt}, CLIPArTT~\cite{hakim2024clipartt}, and WATT~\cite{osowiechi2024watt}. Following WATT, we report the TTA results from the CLIP model to CIFAR~\cite{krizhevsky2009cifar} and various domain adaptation and generalization benchmarks, including VisDA~\cite{peng2018visda}, Office-Home~\cite{venkateswara2017officehome}, PACS~\cite{li2017pacs}, and VLCS~\cite{fang2013vlcs}. As a result, our method achieves competitive performance compared to the previous state-of-the-art WATT, without requiring additional inner-loop training processes or ensemble methods. Furthermore, when WATT is combined with REM, it surpasses the existing results.

\begin{table*}[t]
\centering
\caption{\label{tab:vlm}Classification accuracy (\%) comparison on vision-language model using CLIP ViT/B-16 across different datasets and domains.}
\small
\begin{adjustbox}{width=0.9\linewidth,center=\linewidth}
\begin{tabular}{l|l|cccccccc}
\hline
Dataset&Domain&CLIP&Tent&TPT&CLIPArTT&WATT&\cellcolor{Light!70}REM & \cellcolor{Light!95}REM+WATT\\\hline
&CIFAR-10 &89.25& 92.75& 89.80& 92.61& 91.97&\cellcolor{Light!70}91.76$_{\pm 0.06}$&\cellcolor{Light!95}\bf93.19$_{\pm 0.14}$\\
CIFAR&CIFAR-100 &64.76 &71.73& 67.15& 71.34&72.85&\cellcolor{Light!70}69.15$_{\pm 0.05}$&\cellcolor{Light!95}\bf72.87$_{\pm 0.11}$\\
&Mean &77.01&82.24&78.48&81.98&82.41&\cellcolor{Light!70}80.46&\cellcolor{Light!95}\bf83.03\\\hline

&3D (trainset) &87.16 &87.57 &84.04 &87.58 &87.72&\cellcolor{Light!70}87.45$_{\pm 0.00}$&\cellcolor{Light!95}\bf88.95$_{\pm 0.03}$\\
VisDA-C&YT (valset) &86.61 &86.81&85.90& 86.60& 86.75&\cellcolor{Light!70}\bf86.89$_{\pm 0.01}$&\cellcolor{Light!95}86.84$_{\pm 0.03}$\\
&Mean &86.89& 87.19& 84.97& 87.09 &87.24&\cellcolor{Light!70}87.17&\cellcolor{Light!95}\bf87.90\\\hline

&Art &79.30 &79.26&\bf81.97& 79.34 &80.43&\cellcolor{Light!70}80.17$_{\pm 0.11}$&\cellcolor{Light!95}80.29$_{\pm 0.11}$\\
&Clipart &65.15 &65.64& 67.01& 65.69& 68.26&\cellcolor{Light!70}66.96$_{\pm 0.04}$&\cellcolor{Light!95}\bf68.32$_{\pm 0.09}$\\
Office-Home&Product &87.34& 87.49&\bf89.00&87.35& 88.02&\cellcolor{Light!70}87.77$_{\pm 0.02}$&\cellcolor{Light!95}87.99$_{\pm 0.09}$\\
&RealWorld &89.31 &89.50&89.66&89.29&\bf90.14&\cellcolor{Light!70}\bf90.14$_{\pm 0.01}$&\cellcolor{Light!95}90.08$_{\pm 0.09}$\\
&Mean &80.28 &80.47& \bf81.91 &80.42 &81.71&\cellcolor{Light!70}81.26&\cellcolor{Light!95}81.67\\\hline

&Art &97.44 &97.54&95.10&97.64&97.66&\cellcolor{Light!70}\bf97.71$_{\pm 0.00}$&\cellcolor{Light!95}97.64$_{\pm 0.06}$\\
&Cartoon &97.38 &97.37&91.42& 97.37&97.51&\cellcolor{Light!70}\bf97.53$_{\pm 0.00}$&\cellcolor{Light!95}97.45$_{\pm 0.02}$\\
PACS &Photo &\bf99.58 &\bf99.58& 98.56& \bf99.58& \bf99.58&\cellcolor{Light!70}\bf99.58$_{\pm 0.00}$&\cellcolor{Light!95}\bf99.58$_{\pm 0.00}$\\
&Sketch &86.06 &86.37&87.23& 86.79&89.56&\cellcolor{Light!70}88.35$_{\pm 0.07}$&\cellcolor{Light!95}\bf90.19$_{\pm 0.14}$\\
&Mean &95.12&95.22& 93.08 &95.35 &96.08&\cellcolor{Light!70}95.79&\cellcolor{Light!95}\bf96.22\\\hline

&Caltech101 &\bf99.43 &\bf99.43&97.62&\bf99.43&99.36&\cellcolor{Light!70}99.36$_{\pm 0.00}$&\cellcolor{Light!95}99.39$_{\pm 0.03}$\\
&LabelMe &67.75&67.31&49.77& 67.74& 68.59&\cellcolor{Light!70}68.06$_{\pm 0.12}$&\cellcolor{Light!95}\bf69.26$_{\pm 0.08}$\\
VLCS&SUN09 &71.74 &71.57&71.56& 71.67&  75.16&\cellcolor{Light!70}75.04$_{\pm 0.04}$&\cellcolor{Light!95}\bf75.76$_{\pm 0.10}$\\
&VOC2007 &84.90 &\bf85.10&71.17&84.73&83.24&\cellcolor{Light!70}83.79$_{\pm 0.12}$&\cellcolor{Light!95}83.89$_{\pm 0.20}$\\
&Mean &80.96& 80.85 &72.53 &80.89 &81.59&\cellcolor{Light!70}81.56&\cellcolor{Light!95}\bf82.08\\
\hline
\end{tabular}
\end{adjustbox}
\end{table*}

\section{Experiments on Practical TTA Scenarios}
\noindent\textbf{Computational Time Constraint TTA Scenario.} \cref{time} presents experimental results on ImageNet-3DCC~\cite{kar20223dcc} under the time-constrained protocol~\cite{alfarra2024constraints}. We compare EATA~\cite{niu2022eata} and our proposed method using ViT-B/16. EATA requires 2.41× the time relative to the adaptation speed of $g$ for which $C(g) = 1$, while REM requires 5.10× the time. Therefore, in the episodic scenario, where the model is re-initialized for each domain, REM shows lower performance than EATA due to its relatively slower adaptation. However, in the continual scenario, where domains are learned sequentially without model re-initialization, our method achieves higher performance due to the accumulation of learned knowledge and demonstrates stable adaptation across domains.

\begin{table}[h]
\small
\centering
\caption{Classification error rate (\%) on ImageNet-3DCC under time constraint scenario}
\renewcommand{\arraystretch}{1.2}
\begin{adjustbox}{width=1\linewidth,center=\linewidth}
\begin{tabular}{l|cc|ccc|c|c|ccc|cc|c}
\toprule
Time & \multicolumn{13}{l}{$t\xrightarrow{\hspace*{21.0cm}}$} \\
\toprule
\multirow{2}{*}{Method} & \multicolumn{2}{c|}{Depth of field} & \multicolumn{3}{c|}{Noise} & \multicolumn{1}{c|}{Lighting} & \multicolumn{1}{c|}{Weather} & \multicolumn{3}{c|}{Video} & \multicolumn{2}{c|}{Camera motion} & \multirow{2}{*}{Mean $\downarrow$} \\
 & Near focus & Far focus & Color quant & ISO & Low light& Flash & Fog 3D & Bit err & H.265 ABR & H.265 CRF & XY-mot. blur & Z-mot. blur \\
\midrule
EATA-Episodic & 27.43 & 35.58 & 42.14 & 46.25 & 33.61 & 55.64 & 54.70 & 80.39 & 48.31 & 42.26 & 48.92 & 43.77 & \bf46.58 \\
\rowcolor{Light!70}REM-Episodic & 28.76 & 36.72 & 41.87 & 46.31 & 41.18 & 59.73 & 53.21 & 89.37 & 50.09 & 45.40 & 53.07 & 46.96 & 49.39 \\
\midrule
EATA-Continual & 26.81 & 33.21 & 40.66 & 43.94 & 35.05 & 57.21 & 56.28 & 83.63 & 55.41 & 47.47 & 56.07 & 49.86 & 48.80 \\
\rowcolor{Light!70}REM-Continual & 29.30 & 33.95 & 41.12 & 43.65 & 31.85 & 56.24 & 51.20 & 87.17 & 49.07 & 41.38 & 49.35 & 43.33 & \bf46.47 \\
\bottomrule
\end{tabular}
\end{adjustbox}
\label{time}
\end{table}

\section{Experiments on Different Network Architectures}
The proposed REM leverages the self-attention mechanism of ViT, yet it can be readily extended to other architectures as long as a ranked structure of difficulty can be explicitly defined through chained masking. To demonstrate generality, we introduce two additional variants: one based on feature activation (FA), where the attention map is computed as the average L2-norm of feature vectors across all spatial positions, and another based on Grad-CAM~\cite{selvaraju2017grad} to modulate the masked regions. As shown in \cref{transformer}, FA-based REM consistently improves performance across various Transformer architectures. Additionally, \cref{cnn} demonstrates its applicability to CNNs, where our method achieves notable gains even without the use of self-attention, highlighting the broad utility of the proposed difficulty-aware masking strategy.

\begin{table}[h]
\centering
\small
\vspace{-2mm}
\caption{Mean error rate (\%) on ImageNetC using transformer architectures}
\begin{tabular}{lccccc}
\toprule
Model & Source & Tent & CoTTA & ViDA & \cellcolor{Light!70}REM \\
\midrule
Mobile-ViT-S & 75.28 & 75.61 & 75.72 & 75.27 &\cellcolor{Light!70} 74.28 \\
SwinTransformer-B & 59.26 & 73.17 & 46.84 & 57.84 & \cellcolor{Light!70}46.56 \\
\bottomrule
\end{tabular}
\label{transformer}
\vspace{-3mm}
\end{table}

\begin{table}[h]
\centering
\small
\vspace{-3mm}
\caption{Mean error rate (\%) on ImageNetC using CNN architectures}
\begin{tabular}{lcccccc}
\toprule
Model & CoTTA & EATA & EcoTTA & BECoTTA & \cellcolor{Light!70}REM (FA) & \cellcolor{Light!70}REM (Grad-CAM)\\
\midrule
WideResNet-28 & 16.2 & 18.6 & 16.8 & - & \cellcolor{Light!70}16.9 & \cellcolor{Light!70}16.5 \\
WideResNet-40 & - & 37.1 & 36.4 & 35.5 & \cellcolor{Light!70}34.5 & \cellcolor{Light!70}34.6 \\
\bottomrule
\end{tabular}
\label{cnn}
\end{table}

\section{Calibration Error Analysis}
We investigate the mitigation of model collapse by analyzing the issue of overconfidence through model calibration error. We observe a consistent trend in which the Expected Calibration Error (ECE) tends to increase as the model outperforms the initial source model~\cite{naeini2015ece}. Notably, our method maintains a low ECE while achieving low error rates, emphasizing the practical importance of calibration for reliable and robust adaptation under distribution shifts.

\begin{table}[h]
\centering
\small
\vspace{-2mm}
\caption{Comparison of ECE and error rates on ImageNetC.}
\begin{tabular}{lccccc}
\toprule
ImageNet-C & Source & Tent &SAR & ViDA & \cellcolor{Light!70}REM \\
\midrule
ECE (\%) $\downarrow$ & 5.3 & 12.6 & 10.3 & 14.6 & \cellcolor{Light!70}8.7 \\
Error (\%) $\downarrow$ & 55.8 & 51.0 & 45.2 & 43.4 & \cellcolor{Light!70}39.2 \\
\bottomrule
\end{tabular}
\label{tab:imagenet_c_results}
\end{table}

\section{Comparison with Augmentation-based EM approaches}
Recent augmentation-based EM methods~\cite{marsden2024roid, lee2024cmf} adopt a variety of stochastic augmentations, such as color jitter and random affine transformations, similar to CoTTA and its variants. In contrast, our method introduces an interpretable and sample-specific augmentation scheme based on structured masking. By designing a ranked prediction distribution, we progressively refine the model's predictions while preserving the relative ranking, offering an intuitive and effective adaptation mechanism. ROID and CMT, like EATA, incorporate the Active Sample Criterion (ASC), which omits the backward pass for inaccurately predicted samples by setting their loss to zero. When ASC is applied to our method, it achieves a similar level of computational efficiency. As shown in~\cref{tab:time_error_comparison}, REM with ASC reduces inference time while maintaining competitive error rates. While ASC-based methods enable efficient adaptation by selectively updating on confident samples, our main objective is to leverage the entire set of test samples. Instead of discarding uncertain predictions, REM aims to reduce domain dependency at test time by enforcing a clear intra-image predictive structure, thereby enhancing robustness against unpredictable domain shifts.

\begin{table}[hbt!]
\centering
\small
\vspace{-1mm}
\caption{Comparison of total adaptation time and error rate on ImageNetC.}
\begin{tabular}{lcccc}
\toprule
Method & ROID & CMT & \cellcolor{Light!70}REM (N=1) & \cellcolor{Light!70}REM (N=1, ASC) \\
\midrule
Time & 9m33s & 9m38s & \cellcolor{Light!70}11m47s & \cellcolor{Light!70}9m22s \\
Error (\%) & 41.4 & 40.7 & \cellcolor{Light!70}39.5 & \cellcolor{Light!70}39.7 \\
\bottomrule
\end{tabular}
\label{tab:time_error_comparison}
\end{table}

\section{Failure Case Analysis}
We analyze the discrepancy between the outputs of original and masked images using Total Variation Distance (TVD) in~\cref{tab:tvd_cifar100c}, focusing on two domains where our method achieved significant performance improvements (Gaussian and Shot noise) and two domains where it showed relatively lower performance (Brightness and JPEG). Interestingly, the domains with successful performance gains, e.g., Gaussian and Shot noise, exhibited larger differences in the predicted probability distributions with and without masking. One possible interpretation is that, for relatively easier domains, a small discrepancy between the predicted distributions of the original and masked images may lead to a low loss, which in turn could reduce the adaptation speed. This observation suggests that dynamically adjusting the loss magnitude based on the estimated domain gap may further enhance the adaptation performance. In particular, incorporating an adaptive loss weighting scheme could help balance learning across domains of varying difficulty. We consider this a promising direction and leave its detailed exploration for future work.

\begin{table}[h]
\centering
\small
\caption{Total Variation Distance (TVD) for the first and last 50\% samples under various corruptions on CIFAR100C.}
\begin{tabular}{lcccc}
\toprule
CIFAR100C & \cellcolor{Light!70}Gaussian & \cellcolor{Light!70}Shot & Brightness & Jpeg \\
\midrule
TVD (first 50\%) & \cellcolor{Light!70}$5.54 \pm 1.36$ & \cellcolor{Light!70}$3.44 \pm 1.38$ & $1.69 \pm 0.45$ & $2.90 \pm 0.95$ \\
TVD (last 50\%) & \cellcolor{Light!70}$5.03 \pm 1.38$ & \cellcolor{Light!70}$3.82 \pm 0.69$ & $1.55 \pm 0.41$ & $2.59 \pm 0.78$ \\
\bottomrule
\end{tabular}
\label{tab:tvd_cifar100c}
\end{table}

\section{Hyperparameter Sensitivity Analysis}
\begin{figure}[h]
\begin{minipage}{0.50\textwidth}
    \centering
    \captionof{table}{Mean error rate (\%) for different masking ratios $M_N$ on ImageNetC}
    \small
    \renewcommand{\arraystretch}{1.5}
    \begin{tabular}{p{0.8cm}>{\centering\arraybackslash}p{1.2cm}>{\centering\arraybackslash}p{2.0cm}>{\centering\arraybackslash}p{2.7cm}}
    \toprule
    $M_N$& $M_1$ & $M_2$ & $M_3$ \\
    \midrule
    +5\% & \{0, 5\%\} & \{0, 5\%, 10\%\} & \{0, 5\%, 10\%, 15\%\} \\
    \rowcolor{Light!70}Error & 40.6\% & 39.4\% & 38.9\% \\
    \midrule
    +10\% & \{0, 10\%\} & \{0, 10\%, 20\%\} & \{0, 10\%, 20\%, 30\%\} \\
    \rowcolor{Light!70}Error & 39.7\% & 39.2\% & 39.4\% \\
    \midrule
    +15\% & \{0, 15\%\} & \{0, 15\%, 30\%\} & \{0, 15\%, 30\%, 45\%\} \\
    \rowcolor{Light!70}Error & 39.5\% & 39.4\% & 40.0\% \\
    \bottomrule
    \end{tabular}
    \label{tab:masking}
\end{minipage}
\hfill
\centering
\begin{minipage}{0.43\textwidth}
    \centering
    \includegraphics[width=\linewidth]{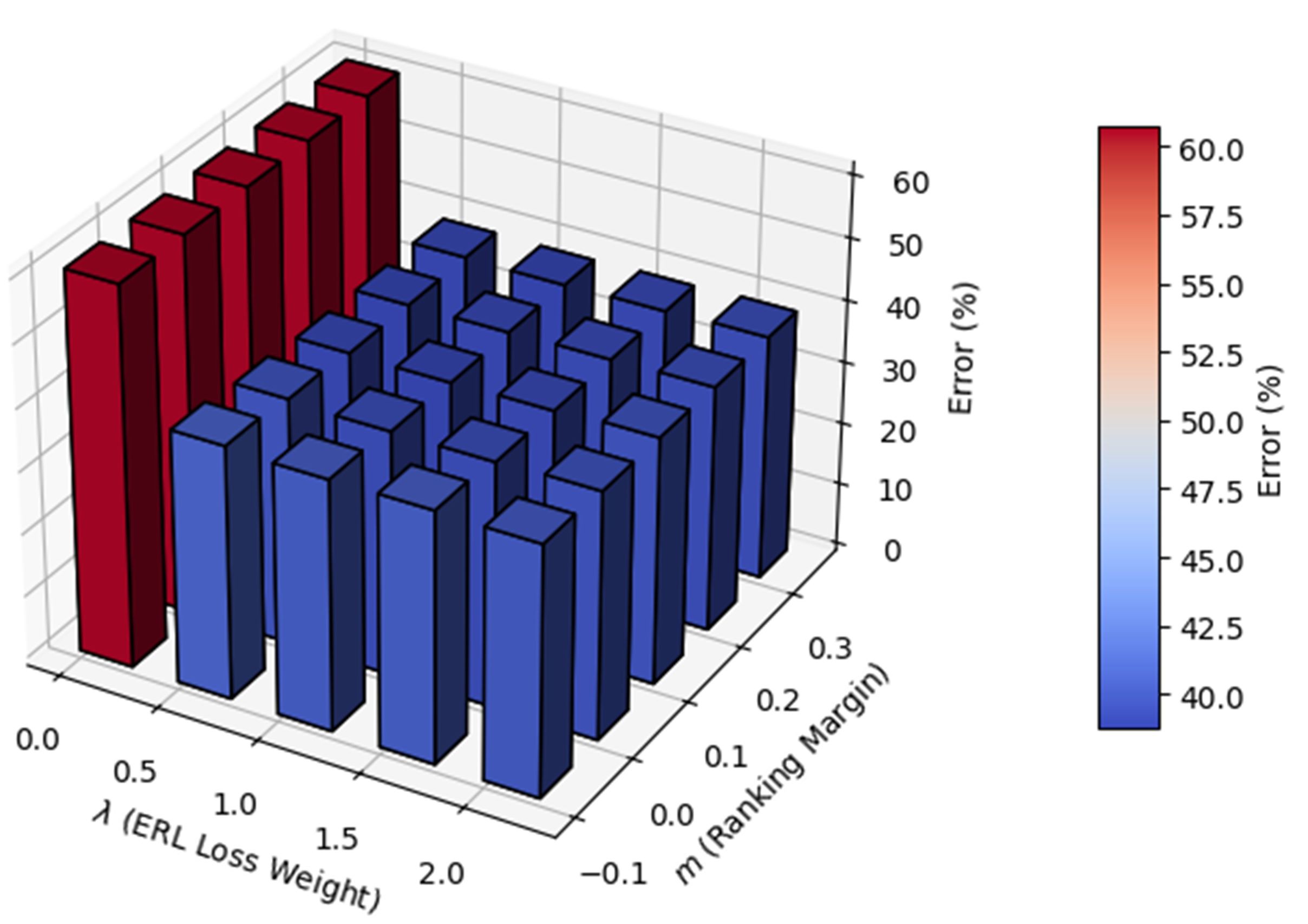}
    \caption{Hyperparameter sensitivity analysis}
    \label{fig:hyperparameter}
\end{minipage}
\end{figure}

REM incorporates three hyperparameters: the masking ratio $M_N$ in~\cref{eq:mcl}, the margin $\mathrm{m}$ in~\cref{eq:erl}, and the weighting coefficient $\lambda$ in~\cref{eq:rem}. We provide an ablation study on the masking ratio in~\cref{tab:masking}. Although the best performance is achieved when $N=3$, we adopt the combination $M_N = \{0, 5\%, 10\%\}$ to strike a balance between computational complexity and accuracy. Moreover,~\cref{fig:hyperparameter} presents the results of a grid search over all hyperparameters, reporting the mean error across all domains in the ImageNet-to-ImageNetC benchmark. The experimental results indicate that our method exhibits low sensitivity to hyperparameter variations. Based on these results, we set $\lambda = 1$ and $\mathrm{m} = 0$ for the CTTA experiments.

\end{document}